# Fast Randomized Singular Value Thresholding for Low-rank Optimization

Tae-Hyun Oh *Student Member, IEEE,* Yasuyuki Matsushita, Yu-Wing Tai *Senior Members, IEEE,*
and In So Kweon *Member, IEEE*

**Abstract**—Rank minimization can be converted into tractable surrogate problems, such as Nuclear Norm Minimization (NNM) and Weighted NNM (WNNM). The problems related to NNM, or WNNM, can be solved iteratively by applying a closed-form proximal operator, called Singular Value Thresholding (SVT), or Weighted SVT, but they suffer from high computational cost of Singular Value Decomposition (SVD) at each iteration. We propose a fast and accurate approximation method for SVT, that we call fast randomized SVT (FRSVT), with which we avoid direct computation of SVD. The key idea is to extract an approximate basis for the range of the matrix from its compressed matrix. Given the basis, we compute partial singular values of the original matrix from the small factored matrix. In addition, by developing a range propagation method, our method further speeds up the extraction of approximate basis at each iteration. Our theoretical analysis shows the relationship between the approximation bound of SVD and its effect to NNM via SVT. Along with the analysis, our empirical results quantitatively and qualitatively show that our approximation rarely harms the convergence of the host algorithms. We assess the efficiency and accuracy of the proposed method on various computer vision problems, *e.g.*, subspace clustering, weather artifact removal, and simultaneous multi-image alignment and rectification.

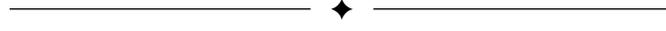

## 1 INTRODUCTION

Minimizing the rank of a matrix can be used as a versatile regularizer to derive a low-rank solution and is needed in many mathematical models in computer vision and machine learning [5], [10], [14], [21], [23], [29], [31], [32], [36], [43], [45]. As rank minimization, $\min \text{rank}(\mathbf{X})$ s.t. $\mathbf{X} \in \mathcal{C}$, is generally an NP-hard problem, it is typically relaxed using the nuclear norm (*i.e.*, $\|\cdot\|_*$, the sum of the singular values), which is a tight convex surrogate for the rank function, *i.e.*, $\text{rank}(\cdot)$.

Nuclear norm minimization (NNM) is expressed as

$$\mathbf{X}^* = \operatorname{argmin}_\mathbf{X} f(\mathbf{X}) + \tau\|\mathbf{X}\|_*, \qquad (1)$$

where $\mathbf{X} \in \mathbb{R}^{m \times n}$, and $\tau > 0$ is a regularization parameter. The function $f(\mathbf{X})$ can be arbitrarily defined depending on the objectives, *e.g.*, $f(\mathbf{X}) = \|\mathbf{O} - \mathbf{X}\|_1$ in robust principal component analysis (RPCA) [4], $f(\mathbf{X}) = \frac{1}{2}\|\mathbf{AX} - \mathbf{B}\|_F^2$ in multivariate regression and multi-class learning [26], $f(\mathbf{X}) = \frac{1}{2}\|\pi_\Psi(\mathbf{O}) - \pi_\Psi(\mathbf{X})\|_F^2$ in matrix completion [4], $f(\mathbf{X}) = \|\mathbf{OX} - \mathbf{O}\|_F^2$ in subspace clustering [21], where $\mathbf{O}$ is measured data, $\|\cdot\|_1$ and $\|\cdot\|_F$ are $l_1$ and Frobenius norms respectively, and $\pi_\Psi(\cdot)$ is an orthogonal projection operator setting $[\pi_\Psi(\mathbf{X})]_{i,j} = [\mathbf{X}]_{i,j}$ for $(i,j) \in \Psi$ and 0 otherwise.

For the purpose of better approximating the rank function $\text{rank}(\cdot)$, weighted nuclear norm minimization (WNNM) [10], [14], [30], which could be non-convex depending on the weights that are used, can be alternatively adopted instead of the standard nuclear norm.

Except for the case that the loss function $f(\mathbf{X})$ is a proximity term, most previous works use first-order optimization approaches, *e.g.*, dual method [8], accelerated proximal gradient [15], augmented Lagrange multiplier (ALM) [18], alternating direction method (ADM) [18], [19], and iteratively reweighted least-squares [24]. In the intermediate step, regardless of regularization, all these approaches have an iterative step to solve a simple NNM, or WNNM, subproblem defined as the following nuclear norm and proximity terms:

**Problem** (**Nuclear norm minimization**). *For $\tau \geq 0$ and $\mathbf{A} \in \mathbb{R}^{m \times n}$,*

$$\mathbf{X}^* = \operatorname{argmin}_\mathbf{X} \tau\|\mathbf{X}\|_* + \frac{1}{2}\|\mathbf{X} - \mathbf{A}\|_F^2, \qquad (2)$$

where optimal $\mathbf{X}^*$ can be obtained by the singular value thresholding operator defined as following.

**Definition 1** (**Singular value thresholding (SVT)**[1] [2]). *The problem (2) has a closed-form solution given by the singular value thresholding operator $\mathbb{S}_\tau(\cdot)$ as*

$$\mathbf{X}^* = \mathbb{S}_\tau(\mathbf{A}) = \mathbf{U}_\mathbf{A}\mathcal{S}_\tau(\mathbf{\Sigma}_\mathbf{A})\mathbf{V}_\mathbf{A}^\top, \qquad (3)$$

*where $\mathcal{S}_\tau(x) = \text{sgn}(x) \cdot \max(|x| - \tau, 0)$ is a soft shrinkage operator [11], and $\mathbf{U}_\mathbf{A}\mathbf{\Sigma}_\mathbf{A}\mathbf{V}_\mathbf{A}^\top$ is the SVD of $\mathbf{A}$.*

The major computational bottleneck of NNM and WNNM problems is the necessity of solving Eq. (2) multiple times, where SVD computation occupies the largest computation cost, *i.e.*, $O(mn\min(m,n))$ for each SVD [9].

This paper proposes a fast SVT technique to accelerate general NNM and WNNM computation. Our method is motivated by the previous study of a randomized SVD proposed by Halko *et al.* [12], and we extend the original general method in several respects for better solving the NNM and WNNM problems that we focus on in this paper. As a result, we propose an algorithm that we call *fast randomized SVT* (FRSVT). We present

---

- T.-H. Oh, Y.-W. Tai, and I.S. Kweon (corresponding author) are with the Department of Electrical Engineering, KAIST, Daejeon, Republic of Korea.
  E-mail: thoh.kaist.ac.kr@gmail.com, yuwing@gmail.com and iskweon77@kaist.ac.kr
- Y. Matsushita is with Osaka University, Japan.
  E-mail: yasumat@ist.osaka-u.ac.jp

[1] A similar result for WNNM can be found in [10], called WSVT.

| | |
|---|---|
| $\pi_\Psi(\cdot)$ | Orthogonal projection operator with a map (index set) $\Psi$ |
| $\mathbb{S}_\tau(\cdot)$ | SVT operator [2] with the parameter $\tau$ |
| $\mathcal{S}_\tau(\cdot)$ | Soft shrinkage operator [11] with the parameter $\tau$ |
| $\sigma_i(\cdot)$ | $i$-th singular value of a matrix |
| $\mathbf{\Omega}$ | A standard Gaussian random matrix |
| $\mathbf{Q}$ | Orthonormal column matrix |
| $\mathbf{O}$ | Observation matrix (Input) |
| $\mathbf{L}, \mathcal{L}$ | Low-rank optimization related matrix or tensor |
| $\mathbf{S}, \mathcal{E}$ | Sparse optimization related matrix or tensor |
| $k$ | Parameter for the target rank |
| $p$ | Parameter for over-sampling rate |
| $l$ | Sampling (predicted) rate ($l = k + p$) |
| $r$ | True rank of the target matrix $\mathbf{A}$ |
| $s$ | Real sampling rate ($s = \min(l, r)$) |
| $\eta$ | Number of the power iteration |
| $b$ | Parameter for the maximum rank bound |
| $\mathbf{v}$ | Parameter vector for the polynomial error bound ($\mathbf{v} = \{k, p\}$ without the power iteration, or $\mathbf{v} = \{k, p, \eta\}$ with it) |

TABLE 1: Summary of notations

the connection between FRSVT and low-rank approximation with both theoretical and empirical analyses, and show the effectiveness of the proposed method via simulations and a few computer vision applications. Table 1 summarizes the notation used in the rest of the paper.

A preliminary version of this paper has appeared in [34]. We extend [34] by analyzing the behavior of the proposed method in both theoretical and empirical aspects as follows. A detailed derivations of error bound on both Frobenius and spectral norms, inexpensive and provable accuracy check, and comparison with a linear-time approximate SVD [6] are newly supplemented in Sec. 3. Based on the analyses, we provide theoretically sound implementation tips and other options that can potentially improve the performance of the proposed algorithm. Additional simulation and experimental results are included in Sec. 4. Specifically, this paper makes the following contributions:

• We develop a successive truncated low-rank decomposition method that can be generally applied to NNM and WNNM problems. Our method achieves high efficiency with rarely degrading the accuracy of the host algorithms.

• By exploiting the proximity of the range space over iterations, our method further accelerate the NNM computation by propagating the estimated basis at the previous iteration to the next one for acceleration. We call this technique *range propagation* (RP).

• We provide a theoretical analysis between the low-rank approximation and our FRSVT method. We show how the low-rank approximation affects to SVT operation as well as effects of the interaction with the power iteration and over-sampling. In addition, we show the empirical stability and behavior of our method with respect to varying parameters.

• We apply FRSVT to various computer vision applications and show the performance gain in comparison with previous methods.

### 1.1 Related Works

Candès *et al*. [4] showed that, under some mild condition, the solution of NNM is equivalent to the solution of rank minimization in conjunction with a sparse outlier model [4]. Inspired by the success of this convex surrogate for rank minimization, low-rank structures have been exploited in various computer vision applications, such as rain removal [5], denoising [10], inpainting [14], motion segmentation by subspace clustering [21], structure from motion [23], background subtraction [29], tag transduction [29], high dynamic range imaging [32], batch image alignment [36], photometric stereo [43], image rectification [45], and nuclear norm regularized learning [26].

While useful, due to the high computational complexity of NNM, specifically, SVD used in the SVT operator, a fast SVT method has always been wanted in both small- and large-scale problems. Liu *et al*. [22] efficiently solve NNM by casting the original convex RPCA problem into a bilinear factorization form. Although this bilinear factorization introduces non-convexity in the objective function like other factorization methods [7], [46], the work achieved significant speed-up. Unlike their method, our method retains the original objective function and approximately optimizes it, and is applicable to general NNM problems once the problems can be led to the form of Eq. (2).

With retaining the advantage of convexity, Liu *et al*. [23] exactly solve RPCA on a small sub-sampled matrix and propagate the seed solution to other parts via $\ell_1$ filtering. Since both Liu *et al*. [22] and Liu *et al*. [23] focus only on RPCA problem but not general NNM problems, a fast and general SVT method is still needed as a tool for NNM to be applied to large-scale problems. Cai *et al*. [3] avoid explicit SVD computation using the dual of SVT. Since their method uses Newton iterations with an inverse matrix, the input matrix needs to be preprocessed by the complete orthogonal decomposition [9] (COD) to ensure a non-singular square matrix. The standard COD consists of twice of QR decompositions with column/row pivoting, which requires $O(mn \min(m, n))$; therefore, the reduction of computation complexity is still limited.

In other thread of works, it has been shown that the exact SVD computation is unnecessary in the inner loops of NNM [18], [26], [29]. Mu *et al*. [29] propose a compressed optimization by random projection. Ma *et al*. [26] solve NNM related problems with a linear-time approximate SVD [6]. However, these methods become occasionally unstable, because the input matrix is directly approximated by sampling or projection, where the original information is impaired, and the randomness leads to unstable and incorrect results. Unlike these methods, our method only approximates the subspace bases to guarantee that most spectrum information is retained.

**Scope of this work** This work mainly focuses on the behavior of the fundamental core module, *i.e*., SVT operator, combined with low-rank approximation. All the provided theoretical analyses in this work are mainly focused on the FRSVT operator *per se*, rather than the overall optimization procedure of low-rank optimization problems. In rank minimization literature, there are approaches [28], [33], [41] that are tightly entangled with the optimization procedure to promote low-rank solutions, and they are beyond SVT like modular operation. This paper does not cover this type of approaches, yet focuses on the SVT operation that is used broadly [5], [10], [14], [21], [26], [32], [36], [43], [45].

## 2 FAST RANDOMIZED SINGULAR VALUE THRESHOLDING (FRSVT)

The basic idea of our method shares the idea of Liu *et al*. [23] and Ma *et al*. [26], in that the solution of Eq. (2) can be found by applying SVT to a small matrix instead of the original large matrix as illustrated in Fig. 1. Instead of sampling columns or rows of a matrix as in [23], [26], our method extracts a small core matrix by finding orthonormal bases with the unitary invariant property. Specifically, since the NNM defined

Fig. 1: Basic idea of our method. Instead of applying SVT to the large original matrix, if we can obtain the same result by applying SVT to small matrix with few additional efforts, the complexity could be significantly drop. As shown in Proposition 1, once a large matrix is decomposed into a core matrix and orthonormal matrices (the left and right thin matrices in the illustration), by applying SVT to the small core matrix we can obtain the same result with the direct SVT computation on the large matrix.

in Eq. (2) consists of unitary invariant norms, the following equality holds:

**Proposition 1.** *Let $\mathbf{A} = \mathbf{QB} \in \mathbb{R}^{m \times n}$, where $\mathbf{Q} \in \mathbb{R}^{m \times n}$ has orthonormal columns. Then,*

$$\mathbb{S}_\tau(\mathbf{A}) = \mathbf{Q}\,\mathbb{S}_\tau(\mathbf{B}), \tag{4}$$

*where $\mathbb{S}_\tau(\cdot)$ is the SVT operator.*

*Proof.* When $m = n$ and $\mathbf{Q}$ is an orthonormal matrix, the equality obviously holds by the unitary invariant property of norms. For $m > n$, we prove the equality for the orthonormal column matrix. Consider an arbitrary matrix $\mathbf{Z} = \mathbf{U_Z \Sigma_Z V_Z}^\top \in \mathbb{R}^{n \times n}$, where $\mathbf{U_Z \Sigma_Z V_Z}^\top$ is $\mathbf{Z}$'s SVD, then

$$\begin{aligned}
\mathbf{Q} \cdot \mathbb{S}_\tau(\mathbf{B}) &= \mathbf{Q} \cdot \operatorname*{argmin}_{\mathbf{Z}} \left( \tau \|\mathbf{Z}\|_* + \frac{1}{2}\|\mathbf{Z} - \mathbf{B}\|_F^2 \right) \\
&= \operatorname*{argmin}_{\mathbf{X}} \left( \tau \|\mathbf{Q}^\top \mathbf{X}\|_* + \frac{1}{2}\|\mathbf{Q}(\mathbf{Z} - \mathbf{B})\|_F^2 \right) \\
&\quad (\text{by letting } \mathbf{QZ} = \mathbf{X}, \text{ and as } \mathbf{Q}^\top \mathbf{Q} = \mathbf{I}, \\
&\quad \text{and the unitary invariant norm property}) \\
&= \operatorname*{argmin}_{\mathbf{X}} \left( \tau \|\mathbf{X}\|_* + \frac{1}{2}\|\mathbf{X} - \mathbf{QB}\|_F^2 \right) \\
&\quad (\text{since } \|\mathbf{Q}^\top \mathbf{X}\|_* = \|\mathbf{Z}\|_* = \|\mathbf{QZ}\|_* = \|\mathbf{X}\|_*) \\
&= \mathbb{S}_\tau(\mathbf{QB}) = \mathbb{S}_\tau(\mathbf{A}).
\end{aligned}$$

□

In general, SVT requires SVD computation, and its complexity[2] is $O(mn^2)$. Based on Proposition 1, we can avoid expensive computation by instead computing SVT on a smaller matrix $\mathbf{B} \in \mathbb{R}^{n \times n}$ when $\mathbf{Q} \in \mathbb{R}^{m \times n}$ is available. Given $\mathbf{Q} \in \mathbb{R}^{m \times k}$ that best approximates $\mathbf{A}$ by a rank-$k$ matrix, the complexity of SVD of $\mathbf{B} \in \mathbb{R}^{k \times n}$ becomes $O(nk^2)$. Therefore, when $k \ll n$, the computation speed can be significantly improved.

Our SVT computation iterates the following two steps: 1) Estimating an orthonormal column matrix $\mathbf{Q}$, and 2) Computing SVD of $\mathbf{B}$ for SVT. For SVT computation, a partial (or truncated) SVD is frequently used to reduce the complexity in many prior arts. Our method similarly finds a rank-$k$ approximation ($k < n$) of the original matrix $\mathbf{A}$ as $\mathbf{A} \approx \hat{\mathbf{A}}_k = \mathbf{QB}$. It saves the computation of the first step as well as the second step, because the size of matrices $\mathbf{Q}$ and $\mathbf{B}$ are reduced to $m \times k$ and $k \times n$, respectively. By exploiting the observation that the major orthonormal $k$ bases evolve slowly over iterations, our method efficiently initializes matrix $\mathbf{Q}$ at each iteration by bypassing expensive random range estimation (Range propagation in Sec. 2.1). In addition, by avoiding direct

[2]Without loss of generality, we assume $m \geq n$ in this paper.

**Algorithm 1** Fast Randomized Singular Value Thresholding (FRSVT) algorithm

1: **Input :** $\mathbf{A} \in \mathbb{R}^{m \times n}, \tau > 0, l = k + p > 0$ and $q \geq 0$. For range propagation, the orthonormal column matrix $\tilde{\mathbf{Q}}$ of the previous iteration.
2:
3: **if not** Range propagation **then**
4: $\quad$ Sample Gaussian random matrix $\mathbf{\Omega} \in \mathbb{R}^{n \times l}$
5: $\quad \mathbf{Y} = \mathbf{A\Omega}$
6: $\quad \mathbf{Q} = \text{QR\_CP}(\mathbf{Y})$
7: **else**
8: $\quad$ Sample Gaussian random matrix $\mathbf{\Omega} \in \mathbb{R}^{n \times p}$
9: $\quad \mathbf{Y} = \mathbf{A\Omega}$
10: $\quad \mathbf{Q_Y} = \text{PartialOrthogonalization}(\tilde{\mathbf{Q}}, \mathbf{Y})$
11: $\quad \mathbf{Q} = [\tilde{\mathbf{Q}}, \mathbf{Q_Y}]$
12: **end if**
13: **repeat**
14: $\quad \mathbf{Q} = \text{QR}(\mathbf{A}\mathbf{A}^\top \mathbf{Q})$
15: **until** $\eta$ times
16: $[\mathbf{H}, \mathbf{C}] = \text{QR}(\mathbf{A}^\top \mathbf{Q})$
17: $[\mathbf{W}, \mathbf{P}] = \text{PolarDecomposition}(\mathbf{C})$
18: $[\mathbf{V}, \mathbf{D}] = \text{EigenDecomposition}(\mathbf{P})$
19: $\mathbb{S}_\tau(\mathbf{A}) = (\mathbf{QV})\,\mathcal{S}_\tau(\mathbf{D})\,(\mathbf{HWV})^\top$
20:
21: **Output :** $\mathbb{S}_\tau(\mathbf{A}), \tilde{\mathbf{Q}} = \mathbf{QV}$

SVD computation, we can further reduce the computation as described in Sec. 2.2. Also, we describe a target rank prediction technique for further improving speed in Sec. 2.3.

### 2.1 Finding Approximate Range

Inspired by Halko *et al.* [12], we first estimate the orthonormal bases $\mathbf{Q} = [\mathbf{q}_1, \cdots, \mathbf{q}_l]$ (where $l \geq k$) such that $\text{span}(\mathbf{Q}) \subseteq \text{Range}(\mathbf{A})$[3] from a matrix compressed by random projection. Intuition of their randomized range finding algorithm is as follows. By multiplying a random vector $\omega_j$, a random linear combination $\mathbf{y}_j$ of the column vectors of $\mathbf{A}$ is generated, which encodes the partial range of $\mathbf{A}$. Suppose $\mathbf{A} = \mathbf{A}_k + \mathbf{E}$, where $\mathbf{A}_k$ is the rank-$k$ projection of $\mathbf{A}$, whose range is the target to be captured, and $\mathbf{E}$ represents small perturbation, then sample vector $\mathbf{y}_j$ can be obtained by

$$\mathbf{y}_j = \mathbf{A}_k \omega_j + \mathbf{E}\omega_j. \tag{5}$$

Even though unwanted $\mathbf{E}$ may be included in $\{\mathbf{y}_j\}$ because the action of $\mathbf{A}_k$ (*i.e.*, magnitudes of spectrum) is larger than $\mathbf{E}$, the range of $\mathbf{A}_k$ is dominant to be captured in $\{\mathbf{y}_j\}$. However, if only $k$ vectors $\{\mathbf{y}_j\}$ are sampled, $\{\mathbf{y}_j\}$ could not span the entire $\text{Range}(\mathbf{A}_k)$. By increasing the sampling rate, most of $\text{Range}(\mathbf{A}_k)$ can be captured; therefore, we oversample $l$ sample vectors, so that $\text{Range}(\mathbf{A}_k)$ can be captured as much as possible.

Given the sample matrix $\mathbf{Y} = [\mathbf{y}_1, \cdots \mathbf{y}_l]$, where $l = k + p$, $\mathbf{Q}$ can be obtained by orthonormalizing $\mathbf{Y}$. When $r = \text{rank}(\mathbf{A}) < l$, $r$ bases are enough to span the entire $\text{Range}(\mathbf{A})$. Indeed, when $\text{rank}(\mathbf{A}) < l$, the range finding algorithm is fairly accurate and close to the exact method as we will see the theoretical analysis in Sec. 3. Thus, we can reduce the dimension of $\mathbf{Q}$ for almost free by estimating the rank and dominant bases by QR decomposition with Column Pivoting (QR-CP) to $\mathbf{Y}$. While Halko *et al.* also proposed an algorithm to adaptively and approximately determine the number of bases by different randomization for sampling, orthogonalization, and re-orthogonalization, our method is based on an exact method with the same complexity using

[3]We only consider the column range space of a matrix for simplicity of explanation, but the row range also can be used.

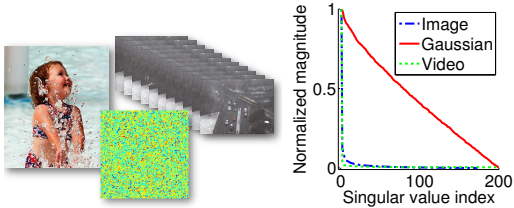

Fig. 2: Illustration of singular value decaying. [Left] Gaussian random, video and image samples. [Right] Decaying graphs of singular values. The *Red*, *Green*, *Blue* lines represent the graphs of a Gaussian random matrix, video, and image, respectively. The spectrum of visual data decays significantly fast.

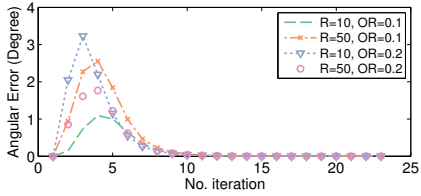

Fig. 3: Angular difference of subspaces between subsequent iterations.

QR-CP, and we adaptively predict the sampling rate in a simpler manner (we will see in Sec. 2.3 or Sec. 3.2). Fortunately, in LAPACK, an efficient QR-CP routine using level-3 BLAS (dgeqp3) is available. Moreover, our method does not require the upper triangle matrix from QR, but only the orthonormal basis $\mathbf{Q}$; therefore, we can avoid extracting the whole triangular matrix but only compute rank and $\mathbf{Q}$ with dgeqp3 and orgqr routines, respectively. After obtaining $\mathbf{Q} \in \mathbb{R}^{m \times s}$, where $s = \min(l, r)$, we can compute a small matrix $\mathbf{B}$ by $\mathbf{B} = \mathbf{Q}^\top \mathbf{A} \in \mathbb{R}^{s \times n}$.

**Range Propagation (RP) for Fast Range Finding** As shown in Fig. 3, we observe that $\mathrm{Range}(\mathbf{A}_{(i)})$ at the $i$-th iteration is similar to the one at the $(i-1)$-th iteration in NNM related problems. This motivates us to use the singular vectors at the $(i-1)$-th step as an initial approximation of range bases $\mathbf{Q}_{(i)}$ at the $i$-th step. A similar observation is exploited in [20]. To capture the change of the range space, we additionally sample $p$ sample vectors $\{\mathbf{y}\}$. We append the sampled $\{\mathbf{y}\}$ to the previous singular vector matrix $\tilde{\mathbf{Q}}_{(i-1)}$ as $\mathbf{Q}_{(i)} = [\tilde{\mathbf{Q}}_{(i-1)}, \mathbf{y}_1, \cdots, \mathbf{y}_p]$, and apply partial orthogonalization only for newly added $\{\mathbf{y}\}$ by a modified Gram-Schmidt procedure [9]. The number of bases can subsequently be reduced by checking the rank with QR-CP in the first step of the power iteration or by a method described in Sec. 3.2.

**Power Iteration** Among the overall process in our algorithm, since the only approximation step is estimation of the orthonormal column matrix $\mathbf{Q}$, the accuracy of our algorithm depends only on this step. In Eq. (5), if the magnitude of the action of $\mathbf{A}_k$ is not dominant against $\mathbf{E}$, the directions of sample vectors are biased and may be affected by portions of $\mathrm{Range}(\mathbf{A}_k)^\perp$. This introduces accuracy degradation to the rest of the process. To resolve this issue, Halko *et al.* [12] used a power iteration scheme, which makes the spectrum difference between $\mathbf{A}_k$ and $\mathbf{E}$ larger by estimating $\mathbf{Q}$ on $(\mathbf{A}\mathbf{A}^\top)^\eta \mathbf{A}$. It improves the chance of better capturing the range of $\mathbf{A}_k$ from $\mathbf{Y} = (\mathbf{A}\mathbf{A}^\top)^\eta \mathbf{A} \mathbf{\Omega}$, while the singular vectors remain unchanged. Halko *et al.* also showed that $\eta = 2$ or $4$ power iterations are sufficient for usual data of interest, and highly accurate range finding can be achieved. As shown in Fig. 2, decay of singular values of visual data is much faster than the one of a Gaussian random matrix. Our empirical tests also show that $\eta = 2$ is sufficient enough, and it is used in all our experiments.

### 2.2 Computing the Singular Values (Vectors)

The NNM problem is now reduced to SVT on a smaller matrix $\mathbf{B}$. In this section, we further reduce the computation time of SVT on $\mathbf{B}$. The SVT operator can be computed by SVD and shrinkage on its singular values. For positive semi-definite matrices, SVD can be more efficiently computed by Eigen decomposition (ED), which is typically faster than SVD in our empirical tests with small matrices. To apply ED to a general matrix, we form a positive semi-definite matrix by the following decomposition:

**Definition 2** (**Polar decomposition** [13]). *Let $\mathbf{X} \in \mathbb{C}^{m \times n}$, $m \geq n$. There exists a matrix $\mathbf{W} \in \mathbb{C}^{m \times n}$ and a unique Hermitian positive semi-definite matrix $\mathbf{P} \in \mathbb{C}^{n \times n}$ such that*

$$\mathbf{X} = \mathbf{W}\mathbf{P}, \qquad \mathbf{W}^* \mathbf{W} = \mathbf{I},$$

*where $\mathbf{I}$ is the identity matrix. If $\mathrm{rank}(\mathbf{X}) = n$, then $\mathbf{P}$ is positive definite and $\mathbf{W}$ is uniquely determined.*

Note that the existence of polar decomposition is equivalent to the existence of SVD.

We use a Newton based polar decomposition suggested by Higham *et al.* [13], which has a quadratic convergence behavior. In our experiment, it converges at a small number of iterations (typically, seven) with various different data, which is consistent with the result of [3], [13]. Due to the requirement of the inverse operator in Newton iterations, it is only applicable to non-singular square matrices. Since $\mathbf{B}^\top \in \mathbb{R}^{n \times s}$ is a full column rank matrix, the non-singular square matrix can be simply obtained from $\mathbf{B}^\top = \mathbf{H}\mathbf{C}$ by QR decomposition, where we call $\mathbf{C} \in \mathbb{R}^{s \times s}$ a *core matrix* that is always non-singular and square. For this step, unlike the procedure in Sec. 2.1, no column pivoting is required.

We sequentially apply the polar decomposition and ED on the core matrix; $\mathbf{C} = \mathbf{W}\mathbf{P} = \mathbf{W}\mathbf{V}\mathbf{D}\mathbf{V}^\top$, where $\mathbf{D}$ and $\mathbf{V}$ are the eigenvalue and eigenvector matrices of $\mathbf{P}$, respectively. Since the matrices $\mathbf{H}$, $\mathbf{W}$, and $\mathbf{V}$ are orthonormal column matrices, the diagonal matrix $\mathbf{D}$ is equivalent to the singular value matrix of $\mathbf{B}$. Finally, $\mathbb{S}_\tau(\mathbf{A})$ can be approximated by

$$\mathbb{S}_\tau(\mathbf{A}) \approx \mathbb{S}_\tau(\hat{\mathbf{A}}_s) = (\mathbf{Q}\mathbf{V})\, \mathcal{S}_\tau(\mathbf{D})\, (\mathbf{H}\mathbf{W}\mathbf{V})^\top. \qquad (6)$$

For the range propagation, the singular vector matrix $\tilde{\mathbf{Q}}$ is stored as $\tilde{\mathbf{Q}} = \mathbf{Q}\mathbf{V}$ or $\mathbf{H}\mathbf{W}\mathbf{V}$ (according to the side of random matrix multiplication). Overall algorithm is summarized in Algorithm 1.

### 2.3 Adaptive Rank Prediction (AP) Heuristic

For SVT, only singular vectors corresponding to the singular values that are greater than a certain threshold are needed, and full SVD is unnecessary. Since the rank of $\mathbf{A}_{(i)}$ is unknown before SVD, predicting its rank can avoid unnecessary computation. We observe that, in many NNM related problems, the rank of $\mathbf{A}_{(i)}$ tends to monotonically increasing or decreasing over iterations, and the rank is stabilized as the number of iteration increases. As we shall see in the theorem of error bound in Sec. 3, over-sampling is always useful to reduce the expected error bound of FRSVT. Thus, optimistically predicting rank allows to achieve both computational efficiency and stability.

The efficiency of our method may be degraded by an excessively high sampling rate. In such a case, we resort to the truncated SVT by upper bounding the target rank. As shown in natural image statistics of Fig. 2, the rank of $\mathbf{A}_{(i)}$ is generally stabilized at low-rank in many computer vision application. Usually, the final accuracy is not harmed, as seen in the successes of the truncated SVD in the NNM related problems [18], [19], [20], [22].

Based on these observations, we define over-sampling rate $p$ as:

$$p_{i+1} = \begin{cases} a, & \text{if } r_i < l_i, \\ \lceil \rho n \rceil, & \text{otherwise,} \end{cases} \quad (7)$$

where $m \geq n$ is assumed, $a$ is a constant (set to $a = 2$), $\rho \in (0, 1]$ is a constant parameter to rapidly follow the real rank $r_i$ (set to $\rho = 0.05$ or less), and $\lceil \cdot \rceil$ denotes the ceil operation. The prediction rule for sampling rate is defined as $l_{i+1} = \min(r_i + p_{i+1}, b)$, where $b = \lceil \gamma n \rceil$ is the maximum bound of sampling rate, $\gamma \in (0, 1]$ is the proportion parameter. The sampling rate $l_i$ can be regarded as the predicted rank at the $i$-th iteration, and $r_i$ is the number of singular values of $\mathbf{A}_{(i)}$ that are larger than the threshold, i.e., the estimated rank of $\mathbb{S}_\tau(\mathbf{A}_{(i)})$. Initially, we set $l_0 = 0.1b$. In the case of the range propagation, the number of columns in $\tilde{\mathbf{Q}}_{(i-1)}$ becomes $r_i$, and $p_i = l_i - r_i$. When $r_i < l_i$, Eq. (7) slightly over-samples, otherwise it optimistically predicts the rank of the next iteration by a larger over-sampling rate. By virtue of low-rankness of visual data shown in Fig. 2, the optimistic rule leads to an accurate estimate.

### 2.4 Fast Random Projection

We can replace the Gaussian random matrix with other less expensive and fast transforms according to the Johnson-Lindenstrauss lemma [27], such as a subsampled random Fourier transform, subsampled random Hadamard transform, or sparse Gaussian random matrix (in Matlab, `sprandn()` function). The cost of the sparse Gaussian random matrix transform depends on the sparsity of the matrix, but it may become less accurate if the matrix is too sparse. The two structured subsampled random matrices can be multiplied with the cost $O(mn \log l)$ by efficient computation methodology (e.g., FFT), while the cost of dense Gaussian matrix multiplication is $O(mnl)$. In our implementation, we simply use dense Gaussian random matrices drawn from standard normal distribution in entry-wise.

### 2.5 Computational Complexity

The step-by-step computational complexities are summarized in Table 2. It does not consider the range propagation procedure for simplicity, but the range propagation only reduces the computational complexity. For example, while the computation of the sample matrix $\mathbf{Y}$ by random projection requires $O(mnl)$, we can reduce it to $O(mn)$ by the range propagation technique.

Since Lines 3 and 4 (i.e., power iteration) in Table 2 iterates $\eta$ times, the overall algorithm in the case of $m < n$ is much more efficient than the other way around. We ensure $\mathbf{A}$ to be $m < n$, if necessary by transposing it, at the beginning of the algorithm. The orthonormalization operation of Line 4 in Table 2 (Line 14 in Alg. 1) is used for enhancing the numerical stability, but in practice it may not be necessary for every step and could be skipped except at the last step. However, in this work, we do not omit it but consistently adopt the orthonormalization every step as described in Alg. 1 for simplicity of evaluation.

| Line No. | Operation | Complexity |
|---|---|---|
| 1 | $\overset{m \times l}{\mathbf{Y}} = \overset{m \times n}{\mathbf{A}} \overset{n \times l}{\mathbf{\Omega}}$ | $O(mnl)$ |
| 2 | $\overset{m \times s}{\mathbf{Q}} = \text{QR\_CP}(\overset{m \times l}{\mathbf{Y}})$ | $O(ms^2)$ |
| 3 | $\overset{m \times s}{\mathbf{Y}} = \overset{m \times n}{\mathbf{A}} \overset{n \times m}{\mathbf{A}^\top} \overset{m \times s}{\mathbf{Q}}$ | $O(mns)$ |
| 4 | $\overset{m \times s}{\mathbf{Q}} = \text{QR}(\overset{m \times s}{\mathbf{Y}})$ | $O(ms^2)$ |
| 5 | $\overset{n \times s}{\mathbf{B}} = \overset{n \times m}{\mathbf{A}^\top} \overset{m \times s}{\mathbf{Q}}$ | $O(ms^2)$ |
| 6 | $[\overset{n \times s}{\mathbf{H}}, \overset{s \times s}{\mathbf{C}}] = \text{QR}(\overset{n \times s}{\mathbf{B}})$ | $O(ns^2)$ |
| 7 | Polar & Eigen Decomposition with $\overset{s \times s}{\mathbf{C}}$ (Lines 17, 18 in Alg. 1) | $O(s^3)$ |
| 8 | Composition $\mathbb{S}_\tau(\mathbf{A})$ (Lines 19 in Alg. 1) | $O(mns)$ |

TABLE 2: Computational complexity of Alg. 1. For simplicity, we do not include the range propagation case, but it can be easily deduced by the above table because most of basic operations remain similar. Incorporating the range propagation obviously requires cheaper computation than the above algorithm.

## 3 ANALYSIS

The proposed method provides an approximate solution of SVT; thus, a natural question is how accurate our algorithm is. This section analyzes this respect by first establishing the relationship between SVT and low-rank approximation. Then, we show that, under what conditions, our method can be expected to be accurate, whereby a few design tips are described for potentially further improving the algorithm. Besides, we show the bound comparison between our method and a representative work, Linear-Time SVD (LTSVD) [6].

### 3.1 Relationship between SVT and low-rank approximation

We can see the relationship between SVT and low-rank approximation by comparing the low-rank approximation gap and SVT operation gap for two different matrices. We begin with reviewing some useful relationships.

The SVT operator $\mathbb{S}_\tau(\cdot)$ is a proximal operator [1]; thus there exists its corresponding dual operator $\mathbb{P}_\tau(\cdot)$ such that $\mathbf{X} = \mathbb{S}_\tau(\mathbf{X}) + \mathbb{P}_\tau(\mathbf{X})$. The dual relationship is valid by the following 2-norm ball Euclidean projection operator [3].

**Definition 3** (2-norm ball Euclidean projection). *For $\tau \geq 0$, consider SVD of $\mathbf{Y} = \mathbf{U_Y} \mathbf{\Sigma_Y} \mathbf{V_Y^\top}$. Then the 2-norm ball Euclidean projection operator is defined by*

$$\mathbb{P}_\tau(\mathbf{Y}) = \mathbf{U_Y} \mathcal{P}_\tau(\mathbf{D_Y}) \mathbf{V_Y^\top},$$

*where $\mathcal{P}_\tau(x) = \min(x, \tau)$ is an element-wise operation.*

Given the 2-norm ball Euclidean projection operator, we have the following lemma.

**Lemma 1.** *The 2-norm ball Euclidean projection operator is the closed-form optimal solution of the following problem:*

$$\mathbb{P}_\tau(\mathbf{Y}) = \min_{\|\mathbf{X}\|_2 \leq \tau} \|\mathbf{Y} - \mathbf{X}\|_F^2.$$

*Proof.* Since the proof is straightforward, we instead provide the idea of the proof. By von Neumann's inequality, the optimal solution of $\mathbf{X}$ should have the same singular vectors with $\mathbf{Y}$. Since the $l_2$-norm and the Frobenius norm are both unitary invariant, the problem is reduced to the diagonal matrix form.

Thus, the projection operator has an explicit form as shown in Definition 3. □

In order to derive a tight error bound, we need contraction inequality, called *pseudo-contraction*[4] [37]:

**Proposition 2 (Pseudo-contraction [37]).** *Let* $\mathbf{A}, \mathbf{B} \in \mathbb{R}^{m \times n}$ *and* $\mathbb{S}_\tau(\cdot)$ *be a proximal operator. Then, the following pseudo-contraction is satisfied.*

$$\|\mathbb{S}_\tau(\mathbf{A}) - \mathbb{S}_\tau(\mathbf{B})\|_F^2$$
$$\leq \|\mathbf{A} - \mathbf{B}\|_F^2 - \|(\mathbb{S}_\tau(\mathbf{A}) - \mathbf{A}) - (\mathbb{S}_\tau(\mathbf{B}) - \mathbf{B})\|_F^2$$
$$= \|\mathbf{A} - \mathbf{B}\|_F^2 - \|\mathbb{P}_\tau(\mathbf{B}) - \mathbb{P}_\tau(\mathbf{A})\|_F^2.$$

The pseudo-contraction holds for all the proximal operators. For deriving a spectral norm based error bound, we derive the spectral pseudo-contraction.

**Proposition 3 (Spectral Pseudo-contraction).** *Let* $\mathbf{A}, \mathbf{B} \in \mathbb{R}^{m \times n}$ *and* $\mathbb{S}_\tau(\cdot)$ *be a proximal operator. There exists a constant* $1 < C \leq 2$ *satisfying*

$$\|\mathbb{S}_\tau(\mathbf{A}) - \mathbb{S}_\tau(\mathbf{B})\|_2 \leq C\|\mathbf{A} - \mathbf{B}\|_2 - \|(\mathbb{S}_\tau(\mathbf{A}) - \mathbf{A}) - (\mathbb{S}_\tau(\mathbf{B}) - \mathbf{B})\|_2. \quad (8)$$

*Proof.* We have the following inequalities for $\mathbb{S}_\tau(\cdot)$ and $\mathbb{P}_\tau(\cdot)$ from non-expansiveness of proximal operator [1]:

$$\|\mathbb{S}_\tau(\mathbf{A}) - \mathbb{S}_\tau(\mathbf{B})\|_2 \leq \|\mathbf{A} - \mathbf{B}\|_2,$$
$$\|\mathbb{P}_\tau(\mathbf{A}) - \mathbb{P}_\tau(\mathbf{B})\|_2 \leq \|\mathbf{A} - \mathbf{B}\|_2.$$

By adding two inequalities, we have

$$\|\mathbb{S}_\tau(\mathbf{A}) - \mathbb{S}_\tau(\mathbf{B})\|_2 \leq 2\|\mathbf{A} - \mathbf{B}\|_2 - \|\mathbb{P}_\tau(\mathbf{A}) - \mathbb{P}_\tau(\mathbf{B})\|_2,$$

and then this implies there exists $C \leq 2$. For the lowest value of $C$, using $\mathbf{A} = \mathbb{S}_\tau(\mathbf{A}) + \mathbb{P}_\tau(\mathbf{A})$ and triangle inequality, we have

$$\|\mathbf{A} - \mathbf{B}\|_2 = \|\mathbb{S}_\tau(\mathbf{A}) + \mathbb{P}_\tau(\mathbf{A}) - (\mathbb{S}_\tau(\mathbf{B}) + \mathbb{P}_\tau(\mathbf{B}))\|_2$$
$$\leq \|\mathbb{S}_\tau(\mathbf{A}) - \mathbb{S}_\tau(\mathbf{B})\|_2 + \|\mathbb{P}_\tau(\mathbf{A}) - \mathbb{P}_\tau(\mathbf{B})\|_2$$
$$\Rightarrow \|\mathbf{A} - \mathbf{B}\|_2 - \|\mathbb{P}_\tau(\mathbf{A}) - \mathbb{P}_\tau(\mathbf{B})\|_2 \leq \|\mathbb{S}_\tau(\mathbf{A}) - \mathbb{S}_\tau(\mathbf{B})\|_2,$$

which implies $C > 1$. We can conclude there exists a constant $C$ such that $1 < C \leq 2$ satisfying Eq. (8). □

Since both pseudo-contractions involve the error term between two projection operators, i.e., $\|\mathbb{P}_\tau(\mathbf{A}) - \mathbb{P}_\tau(\mathbf{B})\|$, the following lemma is useful to see the bound.

**Lemma 2.** *Let* $\hat{\mathbf{A}}_k$ *be a rank-k approximation of* $\mathbf{A}$. *Then,*

$$\min \|\mathbb{P}_\tau(\mathbf{A}) - \mathbb{P}_\tau(\hat{\mathbf{A}}_k)\|_F^2 = \sum_{j>k} \min(\sigma_j(\mathbf{A}), \tau)^2,$$
$$\min \|\mathbb{P}_\tau(\mathbf{A}) - \mathbb{P}_\tau(\hat{\mathbf{A}}_k)\|_2 = \min(\sigma_{k+1}(\mathbf{A}), \tau).$$

*Proof.* Let $\hat{\mathbf{A}}_k = \mathbf{Q}\mathbf{Q}^*\mathbf{A} = \mathbf{P}_\mathbf{Q}\mathbf{A}$, where $\mathbf{Q} \in \mathbb{R}^{m \times k}$ consists of $k$ dominant orthonormal bases, and $\text{Range}(\mathbf{Q}) \simeq \text{Range}(\mathbf{A}_k)$,

[4] Refer to Lemma 3.3 of Pierra *et al.* [37]

where $\mathbf{A}_k$ is the unique and exact rank-$k$ truncated matrix of $\mathbf{A}$, and $\mathbf{P}_\mathbf{Q} = \mathbf{Q}\mathbf{Q}^*$ is an orthogonal projector. Then,

$$\|\mathbb{P}_\tau(\mathbf{A}) - \mathbb{P}_\tau(\hat{\mathbf{A}}_k)\|_F^2$$
$$= \|\mathbb{P}_\tau(\mathbf{A}) - \mathbb{P}_\tau(\mathbf{P}_\mathbf{Q}\mathbf{A})\|_F^2$$
$$= \|\mathbb{P}_\tau(\mathbf{U}_\mathbf{A}\mathbf{\Sigma}_\mathbf{A}\mathbf{V}_\mathbf{A}^\top) - \mathbb{P}_\tau(\mathbf{P}_\mathbf{Q}\mathbf{U}_\mathbf{A}\mathbf{\Sigma}_\mathbf{A}\mathbf{V}_\mathbf{A}^\top)\|_F^2$$
$$= \|\mathbf{U}_\mathbf{A}\mathbb{P}_\tau(\mathbf{\Sigma}_\mathbf{A})\mathbf{V}_\mathbf{A}^\top - \mathbf{P}_\mathbf{Q}\mathbf{U}_\mathbf{A}\mathbb{P}_\tau(\mathbf{\Sigma}_\mathbf{A})\mathbf{V}_\mathbf{A}^\top\|_F^2$$

(by the property of the unitary transform)

$$= \|\mathbf{U}_\mathbf{A}\mathbb{P}_\tau(\mathbf{\Sigma}_\mathbf{A}) - \mathbf{P}_\mathbf{Q}\mathbf{U}_\mathbf{A}\mathbb{P}_\tau(\mathbf{\Sigma}_\mathbf{A})\|_F^2$$

(by the unitary invariant of the norm)

$$= \|(\mathbf{I} - \mathbf{P}_\mathbf{Q})\mathbf{U}_\mathbf{A}\mathbb{P}_\tau(\mathbf{\Sigma}_\mathbf{A})\|_F^2$$
$$= \|\mathbf{P}_\mathbf{Q}^\perp \mathbf{U}_\mathbf{A}\mathbb{P}_\tau(\mathbf{\Sigma}_\mathbf{A})\|_F^2.$$

$\mathbf{P}_\mathbf{Q}^\perp = \mathbf{I} - \mathbf{P}_\mathbf{Q}$ is the orthogonal projector onto the complementary subspace $\text{Range}(\mathbf{P}_\mathbf{Q})^\perp$, which is also close to $\text{Range}(\mathbf{A}_k)^\perp$. Thus, the minimum is achieved only when $\text{Range}(\mathbf{P}_\mathbf{Q}) = \text{Range}(\mathbf{A}_k)$ or $\text{Range}(\mathbf{P}_\mathbf{Q})^\perp = \text{Range}(\mathbf{A}_k)^\perp$ by the well known variational characterization. Therefore,

$$\min \|\mathbf{P}_{\mathbf{A}_k}^\perp \mathbf{U}_\mathbf{A}\mathbb{P}_\tau(\mathbf{\Sigma}_\mathbf{A})\|_F^2 = \sum_{j>k} \min(\sigma_j(\mathbf{A}), \tau)^2.$$

The spectral norm case can be derived similarly. □

Other than the randomization step in sampling $\mathbf{Y}$, our method is based on exact algorithms. Hence, the accuracy is only affected by the randomized range estimation, which computes a rank-$k$ approximation of a matrix. Since there is only one approximation step, the proposed method shares the same theorems with Halko *et al.* [12] as:

**Theorem 1 (Average Frobenius error bounds by randomization [12]).** *Let* $\hat{\mathbf{A}}_k$ *be a rank-k approximation matrix* $\mathbf{A} \in \mathbb{R}^{m \times n}$. *For a target rank* $k \geq 2$ *and an over-sampling parameter* $p \geq 2$, *where* $k + p \leq \min(m, n)$, *draw a standard Gaussian matrix* $\mathbf{\Omega} \in \mathbb{R}^{n \times (k+p)}$. *Then, the theoretical minimum error and the upper bound satisfy*

$$\sum_{j>k} \sigma_j^2(\mathbf{A}) \leq \mathbb{E}\|\mathbf{A} - \hat{\mathbf{A}}_k\|_F^2 \leq \text{poly}(\mathbf{v}) \cdot \sum_{j>k} \sigma_j^2(\mathbf{A}),$$

*where* $\mathbb{E}$ *denotes expectation with respect to the random matrix,* $\text{poly}(\mathbf{v})$ *is a function for* $\mathbf{v} = \{k, p\}$ *without the power iteration or* $\mathbf{v} = \{k, p, \eta\}$ *with the power iteration.*

Halko *et al.* [12] show that the upper bound of the error is close to the theoretical minimum error with high probability in conjunction with some improving techniques (*e.g.*, over-sampling and power iteration), and the bound is rather pessimistic.

In Theorem 1, in the case without power iteration, $\text{poly}(\cdot)$ is defined as:

$$\text{poly}(\mathbf{v} = \{k, p\}) = (1 + k/(p - 1)). \quad (9)$$

The polynomial bound with power iteration in the Frobenius norm representation has no simple form [12]; therefore, instead we observe the error bound with power iteration by referring to the spectral bound in the following Theorem 2 and Corollary 1.

**Theorem 2 (Average spectral error bound by randomization with the power scheme [12]).** *Let* $\hat{\mathbf{A}}_k$ *be a rank-k approximation matrix* $\mathbf{A} \in \mathbb{R}^{m \times n}$. *For a target rank* $k \geq 2$ *and an over-sampling* *parameter $p \geq 2$, where $k + p \leq \min(m, n)$, draw a standard Gaussian matrix $\Omega \in \mathbb{R}^{n \times (k+p)}$. Then, the theoretical minimum error and the upper bound satisfy*

$$\sigma_{k+1} \leq \mathbb{E}\|\mathbf{A} - \hat{\mathbf{A}}_k\|_2 \leq$$
$$\left[\left(1 + \sqrt{\frac{k}{p-1}}\right)\sigma_{k+1}^{2\eta+1} + \frac{e\sqrt{k+p}}{p}\left(\sum_{j>k}\sigma_j^{2(2\eta+1)}\right)^{\frac{1}{2}}\right]^{\frac{1}{2\eta+1}}, \quad (10)$$

*where $\eta$ denotes the number of power iterations.*

**Corollary 1 (Loose average spectral error bound [12]).** *Theorem 2 is bounded as*

$$\mathbb{E}\|\mathbf{A} - \hat{\mathbf{A}}_k\|_2 \leq [\text{Bound of Theorem 2}] \leq \text{poly}(\mathbf{v}) \cdot \sigma_{k+1}, \quad (11)$$

*where* $\text{poly}(\mathbf{v}) = \left[1 + \sqrt{\frac{k}{p-1}} + \frac{e\sqrt{k+p}}{p} \cdot \sqrt{h-k}\right]^{\frac{1}{2\eta+1}}$, $\mathbf{v} = \{k, p, \eta\}$, *and* $h = \min(m, n)$.

Corollary 1 is only useful for understanding the bound behavior rather than identifying the tight bound, because it is far broader than Theorem 2. However, it shows that, as the number of power iterations $\eta$ increases, $\text{poly}(\mathbf{v})$ approaches 1 exponentially fast, and when $\text{poly}(\mathbf{v}) \sim 1$, the spectral error is bounded by the theoretical minimum $\sigma_{k+1}$.

For our method, we can derive the following Frobenius error bound for the approximate SVT (*i.e.*, FRSVT) using Theorem 1, Proposition 2 and Lemma 2.

**Theorem 3 (Average error bound of the approximate SVT).** *Let $\mathbb{S}_\tau(\cdot)$ be the SVT operator and $\hat{\mathbf{A}}_k$ holds Theorem 1. The average error satisfies the following inequality:*

$$\mathbb{E}\|\mathbb{S}_\tau(\mathbf{A}) - \mathbb{S}_\tau(\hat{\mathbf{A}}_k)\|_F^2 \leq \text{poly}(\mathbf{v}) \cdot \left(\sum_{j>k}\sigma_j^2(\mathbf{A})\right) - G(\mathbf{A}),$$

*where $G(\mathbf{A}) = \sum_{j>k}\min(\sigma_j(\mathbf{A}), \tau)^2 \geq 0$.*

*Proof.* By Proposition 2 and Lemma 2,

$$\|\mathbb{S}_\tau(\mathbf{A}) - \mathbb{S}_\tau(\hat{\mathbf{A}}_k)\|_F^2 \leq \|\mathbf{A} - \hat{\mathbf{A}}_k\|_F^2 - \|\mathbb{P}_\tau(\mathbf{A}) - \mathbb{P}_\tau(\hat{\mathbf{A}}_k)\|_F^2$$
$$\leq \|\mathbf{A} - \hat{\mathbf{A}}_k\|_F^2 - \sum_{j>k}\min(\sigma_j(\mathbf{A}), \tau)^2.$$

By taking the expectation of both sides,

$$\mathbb{E}\|\mathbb{S}_\tau(\mathbf{A}) - \mathbb{S}_\tau(\hat{\mathbf{A}}_k)\|_F^2$$
$$\leq \mathbb{E}\|\mathbf{A} - \hat{\mathbf{A}}_k\|_F^2 - \sum_{j>k}\min(\sigma_j(\mathbf{A}), \tau)^2$$
$$\leq \text{poly}(\mathbf{v}) \cdot \left(\sum_{j>k}\sigma_j^2(\mathbf{A})\right) - \sum_{j>k}\min(\sigma_j(\mathbf{A}), \tau)^2,$$
(by Theorem 1)
$$= \text{poly}(\mathbf{v}) \cdot \left(\sum_{j>k}\sigma_j^2(\mathbf{A})\right) - G(\mathbf{A})$$

Obviously, $G(\mathbf{A}) \geq 0$ by definition. □

As aforementioned, $\text{poly}(\mathbf{v})$ has a simple form of Eq. (9) only when the power iteration is not taken into account. For the case with power iteration, it is useful to see the spectral error bound of the approximate SVT, which is our main result.

**Theorem 4 (Average spectral error bound of the approximate SVT with the power scheme).** *Let $\mathbb{S}_\tau(\cdot)$ be the SVT operator and $\hat{\mathbf{A}}_k$ holds Theorem 2. The average spectral error satisfies the following inequality:*

$$\mathbb{E}\|\mathbb{S}_\tau(\mathbf{A}) - \mathbb{S}_\tau(\hat{\mathbf{A}}_k)\|_2 \leq$$
$$C\left[\left(1 + \sqrt{\frac{k}{p-1}}\right)\sigma_{k+1}^{2\eta+1} + \frac{e\sqrt{k+p}}{p}\left(\sum_{j>k}\sigma_j^{2(2\eta+1)}\right)^{\frac{1}{2}}\right]^{\frac{1}{2\eta+1}} - G'(\mathbf{A}),$$
(12)

*where $1 < C \leq 2$, and $G'(\mathbf{A}) = \min(\sigma_{k+1}(\mathbf{A}), \tau) \geq 0$.*

*Proof.* We can simply derive Eq. (12) in a similar manner to Theorem 3, but instead by using Theorem 2 and Proposition 3. □

**Corollary 2 (Loose average spectral error bound of the approximate SVT with power scheme).** *Theorem 4 is bounded as*

$$\mathbb{E}\|\mathbb{S}_\tau(\mathbf{A}) - \mathbb{S}_\tau(\hat{\mathbf{A}}_k)\|_2 \leq [\text{Bound of Theorem 4}]$$
$$\leq C \cdot \text{poly}(\mathbf{v}) \cdot \sigma_{k+1} - G'(\mathbf{A}),$$

*where* $\mathbf{v} = \{k, p, \eta\}$, $\text{poly}(\mathbf{v}) = \left[1 + \sqrt{\frac{k}{p-1}} + \frac{e\sqrt{k+p}}{p} \cdot \sqrt{h-k}\right]^{\frac{1}{2\eta+1}}$, $h = \min(m, n)$, $1 < C \leq 2$ *and* $G'(\mathbf{A}) = \min(\sigma_{k+1}(\mathbf{A}), \tau) \geq 0$.

*Proof.* Since $\sigma_j \geq \sigma_{j+1}$ for any $j$, the term $\left(\sum_{j>k}\sigma_j^{2(2\eta+1)}\right)^{\frac{1}{2}}$ in Eq. (12) is upper-bounded by $\sqrt{h-k} \cdot \sigma_{k+1}^{2\eta+1}$. Then, re-expressing the inequality leads the conclusion. □

Consequently, both of Theorems 3 and 4 assert that the bounds of the approximate SVT are tighter than the respective errors by rank-$k$ approximation in Theorems 1 and 2 on average. Thus, we have the following properties from Theorems 3 and 4:

- Once $\sigma_{k+1}(\mathbf{A})$ approaches a small value, then $\|\mathbb{S}_\tau(\mathbf{A}) - \mathbb{S}_\tau(\hat{\mathbf{A}}_k)\|$ may approach close to zero, *i.e.*, which indicates the approximation is almost exact, roughly when $\text{rank}(\mathbf{A}) \leq k$. Detailed analyses of how we can exploit this property in the algorithm are described in Sec. 3.2.
- Fortunately, the fact that the bound becomes rapidly tighter as $p$ or $\eta$ increases in both practice and theory remains consistent with the result of Halko *et al.* [12].
- The bound is independent of the size of the input matrix.

### 3.2 Residual Singular Value Check without Singular Value Computation

In the previous section, we state that our algorithm may produce almost exact results when $k \geq \text{rank}(\mathbf{A})$, *i.e.*, $\sigma_{k+1} = 0$. We may have a broader condition for ensuring accurate estimates, because the bounds in Theorems 3 and 4 are tighter than those for low-rank approximation in Theorems 1 and 2, respectively. In this section, we explain the condition in which the proposed method can be expected to be accurate, and how the condition can be verified in a computationally inexpensive manner.

Given the target rank $k$ and an approximated matrix $\hat{\mathbf{A}}$ by the range approximation, if $\sigma_k(\hat{\mathbf{A}}) \leq \tau$, then no accuracy improvement is possible. This is because all the remaining singular values $\{\sigma_j(\hat{\mathbf{A}})\}_{j \geq k}$ are zeroed out by thresholding operation with $\tau$, and these does not affect the solution any further. Unless $\sigma_k(\hat{\mathbf{A}}) \leq \tau$, we can additionally estimate a few more bases as described in Sec. 2.1, and we may check the condition again until a desired accuracy is achieved. This procedure can be adopted with Gram-Schmidt procedure [9] in Line 10 of Alg. 1 when an accurate estimate is required.

While the complexity of the single additional basis estimation only affects the factor of $s$ in Table 2 by $s \leftarrow s + 1$, which still preserves the original complexity, checking $\sigma_k(\hat{\mathbf{A}})$ requires additional computation because it is unavailable until the end of the FRSVT algorithm. Instead of directly estimating $\sigma_k(\hat{\mathbf{A}})$, we can utilize the following useful lemma.

**Lemma 3** ( [42]). *Let $\mathbf{A} \in \mathbb{R}^{m \times n}$, and let $q$ be a positive integer, $\alpha$ be a real number greater than 1. By drawing an independent family $\{\omega^{(i)} : i = 1, 2, ..., q\}$ of standard Gaussian vectors, then,*

$$\|\mathbf{A}\|_2 \leq \alpha \sqrt{\tfrac{2}{\pi}} \max_{i=1,...q} \|\mathbf{A}\omega^{(i)}\|_2 \qquad (13)$$

*with probability at least $1 - \alpha^{-q}$.*

The lemma illustrates the relationship between the true largest singular value and the estimate by Gaussian random projection. From this, we derive the following proposition.

**Proposition 4** (Residual singular value estimate). *Let $\mathbf{Q}$ be any rank-$k$ orthonormal column matrix. Given $\mathbf{Q}$, the $(k+1)$-th singular value is bounded by $\sigma_{k+1}(\mathbf{A}) \leq \|(\mathbf{I} - \mathbf{Q}\mathbf{Q}^*)\mathbf{A}\|_2$. Then, drawing a standard Gaussian vector $\omega$,*

$$\|(\mathbf{I} - \mathbf{Q}\mathbf{Q}^*)\mathbf{A}\|_2 \leq \alpha \sqrt{\tfrac{2}{\pi}} \|(\mathbf{I} - \mathbf{Q}\mathbf{Q}^*)\mathbf{A}\omega\|_2, \qquad (14)$$

*with probability at least $1 - \alpha^{-1}$.*

*Proof.* From the well known variational characterization of singular values, we can obtain the minimum of $\|(\mathbf{I} - \mathbf{Q}\mathbf{Q}^*)\mathbf{A}\|_2$ with $\mathbf{Q}^*\mathbf{Q} = \mathbf{I}$ (orthonormal column matrix) iff $\mathbf{Q} = \mathbf{U}_k$, where $\mathbf{U}_k$ is the singular vector matrix of $\mathbf{A}$ corresponding to the top-$k$ largest singular values. Thus,

$$\|(\mathbf{I} - \mathbf{Q}\mathbf{Q}^*)\mathbf{A}\|_2 \geq \|(\mathbf{I} - \mathbf{U}\mathbf{U}^*)\mathbf{A}\|_2 = \sigma_{k+1}(\mathbf{A}).$$

By drawing a single standard Gaussian random vector $\omega$, the left side of the above inequality can be further upper-bounded by Lemma 3 as Eq. (14) with probability at least $1 - \alpha^{-1}$. This concludes the proof. □

Since $\mathbf{y} = \mathbf{A}\omega$ in Eq. (14) is the same by-product with the over-sampling during range propagation procedure in Sec. 2.1, we can check the provable upper bound estimate of $(k+1)$-th singular value $\sigma_{k+1}(\hat{\mathbf{A}}) (\leq \sigma_{k+1}(\mathbf{A}))$ by $\varsigma_{k+1} = \alpha\sqrt{\tfrac{2}{\pi}}\|\mathbf{y} - \mathbf{Q}\mathbf{Q}^*\mathbf{y}\|_2$ with high probability (say 95% with $\alpha = 20$), where the residual vector computation $\mathbf{y} - \mathbf{Q}\mathbf{Q}^*\mathbf{y}$ is also a by-product of the Gram-Schmidt procedure. Notice that all the computation for estimating the quantity $\varsigma_{k+1}$ has been already performed in the original FRSVT procedure; therefore, no additional computation is required except for a scaling with $\alpha\sqrt{\tfrac{2}{\pi}}$, which is almost negligible. Thus, by checking $\varsigma_{k+1} \leq \tau$, we are able to avoid unnecessary additional sampling.

### 3.3 Bound Comparison with Linear-Time SVD

We show the error bound comparison with a relevant method by showing the approximation error bound of Drineas *et al.* [6].

**Theorem 5** (Average error bound of LTSVD [6]). *Let $\mathbf{A} \in \mathbb{R}^{m \times n}$ and $\mathbf{H}_k$ be a matrix created by the LTSVD algorithm [6] by sampling $c$ columns of $\mathbf{A}$ with probabilities $\{p_i\}_{i=1}^n$ such that $p_i \geq \beta |\mathbf{A}^{(i)}|^2/\|\mathbf{A}\|_F^2$ for some positive $\beta \leq 1$ and $\epsilon > 0$. If $c \geq 4k/\beta\epsilon^2$, then the average Frobenius error is bounded by*

$$\mathbb{E}\|\mathbf{A} - \mathbf{H}_k\mathbf{H}_k^\top\mathbf{A}\|_F^2 = \mathbb{E}\|\mathbf{A} - \hat{\mathbf{A}}_k\|_F^2 \leq \|\mathbf{A} - \mathbf{A}_k\|_F^2 + \epsilon\|\mathbf{A}\|_F^2.$$

*In addition, the average spectral error is bounded by*

$$\mathbb{E}\|\mathbf{A} - \mathbf{H}_k\mathbf{H}_k^\top\mathbf{A}\|_2 = \mathbb{E}\|\mathbf{A} - \hat{\mathbf{A}}_k\|_2 \leq \sqrt{\|\mathbf{A} - \mathbf{A}_k\|_2^2 + \epsilon\|\mathbf{A}\|_F^2}.$$

For a fair comparison, we set the same parameters when comparing the low-rank approximation bounds of Drineas *et al.* and Halko *et al.* [12], *i.e.*, the over-sampling rate to be the same as the specified target rank $k$.

**Corollary 3.** *Suppose that the sampling rate $c$ is set twice larger than the target rank $k$, i.e., $c = 2k$, then $\epsilon \geq \sqrt{2}$, and LTSVD also holds the following bounds:*

$$\mathbb{E}\|\mathbf{A} - \hat{\mathbf{A}}_k\|_F^2 \leq \sum_{j>k} \sigma_j^2(\mathbf{A}) + \sqrt{2}\sum_{j=1}^{\min(m,n)} \sigma_j^2(\mathbf{A}), \quad (15)$$

$$\mathbb{E}\|\mathbf{A} - \hat{\mathbf{A}}_k\|_2 \leq \left(\sigma_{k+1}^2(\mathbf{A}) + \sqrt{2}\sum_{j=1}^{\min(m,n)} \sigma_j^2(\mathbf{A})\right)^{\frac{1}{2}}. \quad (16)$$

*Proof.* From Theorem 5, we have $\beta \leq 1$, $\epsilon > 0$ and $c \geq 4k/\beta\epsilon^2$. Hence, $\epsilon \geq 2\sqrt{\tfrac{k}{\beta c}}$. The lowest bound can be achieved at $\beta = 1$; therefore, if $c = 2k$ is assumed, $\epsilon \geq 2\sqrt{\tfrac{k}{c}} = \sqrt{2}$. In addition, we can obtain Eqs. 15 and 16 from the following bounds by using the above result with Theorem 5.

$$\mathbb{E}\|\mathbf{A} - \hat{\mathbf{A}}_k\|_F^2 \leq \|\mathbf{A} - \mathbf{A}_k\|_F^2 + \sqrt{2}\|\mathbf{A}\|_F^2.$$

The average spectral bound can be derived analogously. □

From Corollary 3 and Theorem 1, we obtain the following conclusion.

**Proposition 5.** *Suppose the target rank $k \geq 4$ and the over-sampling rate to be $k$ (for Halko et al. [12], $p = k$, and for LTSVD, $c = 2k$), then on average Halko et al. [12] without power iteration has a tighter bound than LTSVD.*

*Proof.* We compare the right bound terms of Theorem 1 and Corollary 3 in Eq. (15).

$$[\text{LTSVD}] \quad \sum_{j>k} \sigma_j^2(\mathbf{A}) + \sqrt{2}\sum_{j=1}^{\min(m,n)} \sigma_j^2(\mathbf{A}). \quad (17)$$
$$[\text{Halko et al.}] \qquad \text{poly}(\mathbf{v}) \cdot \sum_{j>k} \sigma_j^2(\mathbf{A}). \quad (18)$$

Suppose that Eq. (17) < Eq. (18).

$$\sum_{j>k} \sigma_j^2(\mathbf{A}) + \sqrt{2}\sum_{j=1}^{\min(m,n)} \sigma_j^2(\mathbf{A}) < \text{poly}(\mathbf{v}) \cdot \sum_{j>k} \sigma_j^2(\mathbf{A})$$

$$\Rightarrow \quad -\tfrac{k}{k-1}\sum_{j>k} \sigma_j^2(\mathbf{A}) + \sqrt{2}\sum_{j=1}^{\min(m,n)} \sigma_j^2(\mathbf{A}) < 0$$

$$\Rightarrow \quad \sqrt{2}\sum_{1 \leq j \leq k} \sigma_j^2(\mathbf{A}) + C(k) \cdot \sum_{j>k} \sigma_j^2(\mathbf{A}) < 0 \quad (19)$$

$$\left(C(k) = \sqrt{2} - \tfrac{k}{k-1}\right)$$

For $k \geq 4$, the minimum of $C(k)$ is achieved at $k = 4$ as $C(4) = \sqrt{2} - \tfrac{4}{3} > 0$. Thus, since $C(k) > 0$ for $k \geq 4$, the left term of Eq. (19) is positive for $k \geq 4$. This contradicts the assumption that Eq. (17) < Eq. (18), which concludes the proof. □

For simplicity, we make a reasonable assumption, the target rank $k \geq 4$. Since the bound comparison is performed without the power iteration scheme, Halko *et al.* [12] may have a loose bound. However, as shown in Proposition 5, Halko *et al.* [12] still result in a better approximation bound than LTSVD even under a non-ideal condition. Accompanied with the power iteration, we may be able to derive broader conditions that Halko *et al.* [12] perform better than LTSVD in a similar manner to the proof process of Proposition 5. We would like note that LTSVD is still fascinating in that it does not require dense matrix multiplications for Gaussian matrix sketching used in Halko *et al.* [12] by computing the approximate leveraging score $\{p_i\}$ which is computationally less expensive.

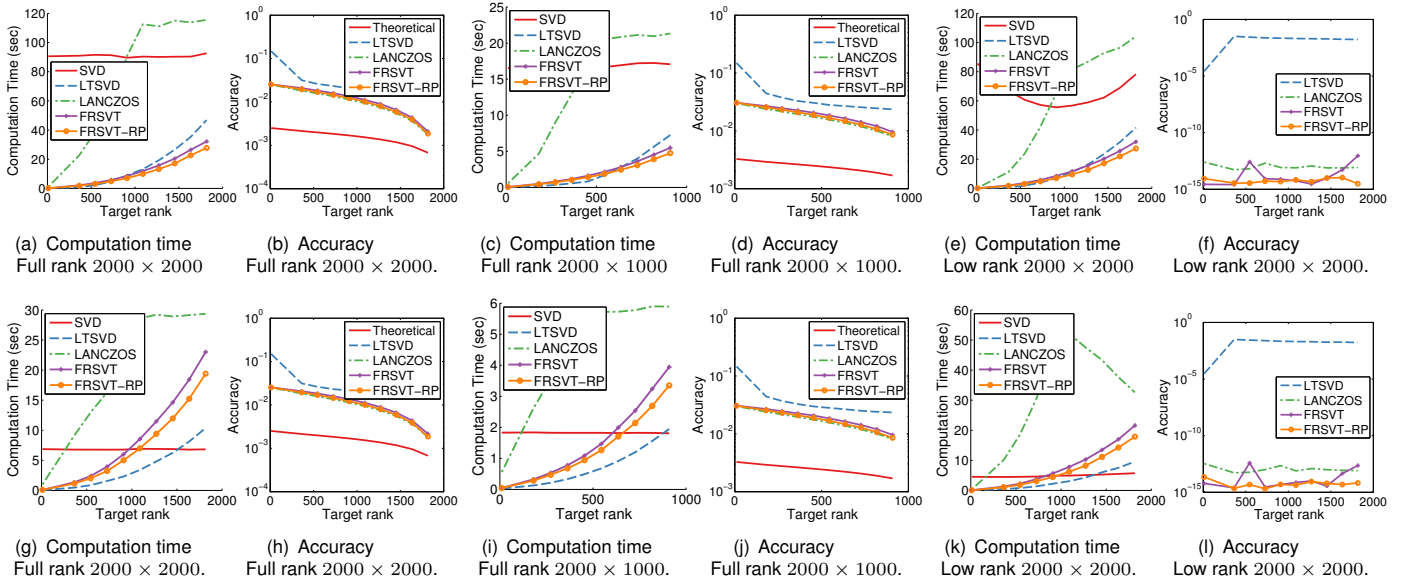

Fig. 4: SVT comparisons among SVD methods. [Top] Experiments on Matlab 2010a. [Bottom] Experiments on Matlab 2014a. Accuracy is measured by normalized root mean squared error (NRMSE), $\|\mathbf{A}^* - \hat{\mathbf{A}}_k\|_F / \|\mathbf{A}^*\|_F$, where $\mathbf{A}^* = \mathbb{S}_\tau(\mathbf{D}_{GT})$, $\hat{\mathbf{A}}_k = \mathbb{S}_\tau(\mathbf{D}_k)$, $\mathbf{D}_{GT}$ is the input data, and $\mathbf{D}_k$ is approximated by each method. For the low-rank matrix test, we generate input data matrices whose rank correspond to the target rank by multiplying two Gaussian random matrices with $m \times r$ and $r \times n$, while in other tests full rank matrices are used. In (b,d,h,j), the theoretical minimum error bound by rank truncation in $\mathbb{S}_\tau(\mathbf{D}_k)$ is provided for guidance, and it is defined as $\sum_{j>k} \sigma_i(\mathbf{A}^*)$, where $\sigma_i(\mathbf{A}^*)$ is computed by Matlab built-in SVD. For the low-rank case in (f,l), the theoretical minimum error is zero, so we omit the theoretical bound.

## 4 EXPERIMENTAL RESULTS

In this section, we first evaluate the efficiency of the proposed method in comparison with other methods using simulation data. The evaluation is conducted by examining the performance of a single SVT computation and also by RPCA computation, which is arguably the most relevant NNM problem in computer vision today. We then show NNM applications in computer vision using real-world data; subspace clustering, semi-online weather artifact removal, and simultaneous multi-image alignment and rectification. All the experiments are conducted on a PC with Intel i7-3.4GHz and 16GB RAM. The same shared parameters were used among different algorithms.

### 4.1 Evaluation using Simulation Data

We quantitatively evaluate our method in comparison to other methods with synthetic matrices sampled from a standard Gaussian distribution. Computational times are evaluated using Matlab 2010a and 2014a 64bits. Since the recent Matlab has been intensively optimized on Intel CPU (mainly due to improvement of Intel MKL), the computation efficiency has been noticeably improved. Therefore, it is worth reporting the performance difference on these two versions of Matlab, because most related works have been assessed with Matlab older than 2011a [3], [20], [22], [23], [29]. For a fair comparison, we turn off multi-threading functions including `maxNumCompThreads(1)` in Matlab. We describe the implementation details of each method in Table 3.

**Single SVT test**   Figure 4 compares speed and accuracy of SVT computation using various SVT methods, such as Matlab built-in SVD[5] (baseline), Lanczos [17], FSVT [3], LTSVD [6] and our FRSVT (with / without range propagation (RP hereafter)). All the implementation details are summarized in Table 3. Except for the baseline SVD, the others produce the truncated SVD of the input matrix, and it is used for SVT computation.

We test using Gaussian random matrix drawn from $N(0,1)$ for the operation $\mathbb{S}_\tau[\cdot]$ with $\tau = 1/\sqrt{\|\mathbf{A}\|_2}$. For a rank-$k$ approximation, we compute rank-$(k+p)$ approximations by setting the over-sampling rate $p$ to 2 for Lanczos, $k$ for LTSVD, 5 for FRSVT.[6] We apply the power iteration twice, i.e., $\eta = 2$. While LTSVD in Fig. 4-(g), (i), (l) is faster than ours, the approximation error is significantly higher than any other truncated SVDs, and it results in slower convergence in RPCA as we will see in the following test.

**Robust PCA test**   To observe the convergence behavior of SVT methods in RPCA [4], we compare our method with various SVT methods, such as SVD, LTSVD, BLWS [20], FSVT, RSVD[7], using an inexact augmented Lagrange multiplier method [18] (iALM, or called alternating directional multiplier method), and other variants, such as Mu et al. [29][8], Active Subspace [22], $L_1$ filtering [23] in Table 5 for resultant performance. Also, the continuous convergence behaviors are shown in Figs. 5 (over the number of iteration) and 6 (over computation times).[9] We generate data by following [35] with $10\%$ rank ratio and $5\%$ corruption ratio. We set the trade-off parameter $\lambda = \frac{1}{\sqrt{\max(m,n)}}$ as suggested by Candès et al. [4] for all the methods. All other parameters for FRSVT are same as mentioned in Sec. 2 and with Table 4. We apply the proposed adaptive rank prediction described in Sec. 2.3 to LTSVD, RSVD and our FRSVT. LTSVD shows convergence

---

[5] We apply either the economic size or full SVD depending on the matrix shape and report the faster one, but we denote either as econSVD.

[6] The rationals behind the used parameters are that Lanczos is an almost exact method, LTSVD produces quite low accuracy despite high over-sampling rate, and FRSVT is an approximate method fairly more accurate than LTSVD even with low over-sampling rate.

[7] As noted in Table 3, this implementation involves 2-side random projection and respective power iterations. By this, we can see the effects of the different implementation of [12].

[8] For Mu et al. [29], we could not reproduce their result. However, we include the results for thoroughness. The results are consistent with the early reported results in Liu et al. [23].

[9] The convergence of $L_1$ filtering is not compared, because the algorithm solves RPCA by dividing several block matrices of which convergence is not comparable to other methods.

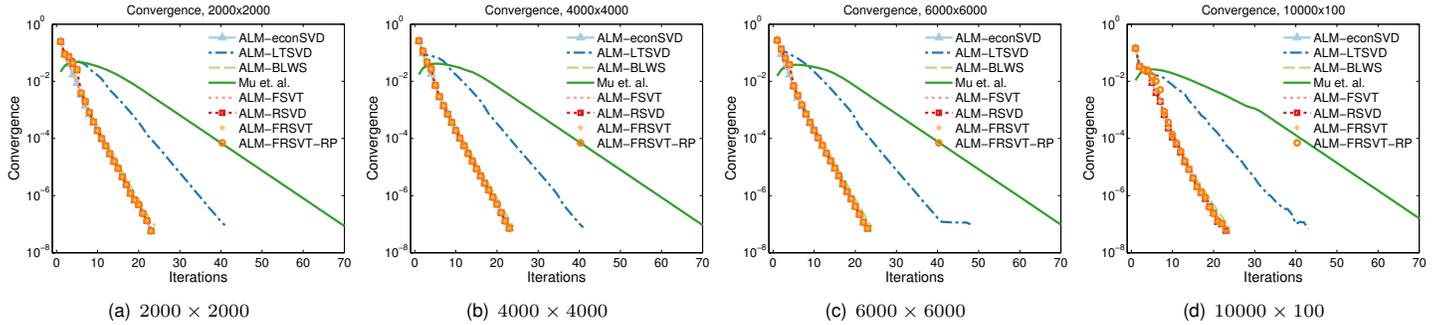

(a) $2000 \times 2000$   (b) $4000 \times 4000$   (c) $6000 \times 6000$   (d) $10000 \times 100$

Fig. 5: RPCA comparisons. X-axis: Number of iterations, Y-axis: Stopping criterion. Except Mu *et al*. is based on the exact ALM method, all the other methods are based on the inexact ALM.

| | |
|---|---|
| SVD$_{\text{Matlab 2010a}}$ | Built-in SVD(·, 'econ') is based on Intel MKL 10.2.2 and LAPACK 3.2.1. |
| SVD$_{\text{Matlab 2014a}}$ | Built-in SVD(·, 'econ') is based on LAPACK 3.4.1 in Intel MKL 11.0.5. |
| LTSVD* | The Matlab implementation by Ma *et al*. [26] is used. |
| Lanczos | The implementation optimized by Lin *et al*. [17] is used. It is implemented by Matlab and Mex to use LAPACK (ver. 3.4.1 in Intel MKL 11.0.5 is used). |
| BLWS | The Matlab implementation by Lin *et al*. [20] is used. |
| Mu *et al*. | The Matlab implementation provided by Mu *et al*. [29] is used. All the parameters are directly used as suggested. |
| Active Subspace | The Matlab implementation provided by Liu *et al*. [22] is used. All the parameters are directly used as suggested. |
| $L_1$ filtering | The Matlab implementation provided by Liu *et al*. [23] is used. All the parameters are directly used as suggested. |
| FSVT *** | We implement FSVT by Matlab and Mex based on Intel MKL 11.0.5, and the speed-up gain is checked and consistent with the reported results in Cai *et al*. [3]. All the parameters are directly used as suggested. |
| RSVD *** | We implement RSVD [12] based on LAPACK 3.4.1 in Intel MKL 11.0.5. by referring to RedSVD**. |
| FRSVT *** | Our Matlab implementation based on Intel MKL 11.0.5. Partially, we implement Mex functions for QR–CP (dgeqp3) and Polar Decomposition with BLAS and LAPACK 3.4.1 routines. |

* LTSVD: In the implementation of [26], the leverage score estimation suggested by the original paper [6] is omitted, which is required for the importance sampling. So we implement it with $O(mn)$ complexity.
** RedSVD (http://code.google.com/p/redsvd/): Since RedSVD is implemented based on Eigen 3.0-b1, without the power iteration but with 2-side random projection, so the original implementation is not fast and inaccurate. Thus, we re-implement RSVD [12] by referring to RedSVD with 2-side power iteration.
*** For FSVT, RSVD and FRSVT implementation, we utilize Armadillo library [38] as a high level wrapper for LAPACK and BLAS of Intel MKL in our Mex implementation part. In the Mex functions of Lanczos, Lin *et al*. [17] directly call LAPACK routines without any external library.

TABLE 3: Lists of comparison methods and details for SVT tests. All the parallelization techniques are turned off for fair comparison.

| iALM$_{\text{Methods}}$ | econSVD | LTSVD | BLWS | FSVT | RSVD | FRSVT | FRSVT-RP |
|---|---|---|---|---|---|---|---|
| $p$ | – | +5 | – | – | +2 | +2 | +2 |
| AP | × | ○ | ○ | × | ○ | ○ | ○ |
| PI | – | – | – | – | 2 | 2 | 2 |

TABLE 4: Parameter settings for convergence evaluation of RPCA. ($p$: Over-sampling rate, AP: Adaptive rank prediction, PI: The number of power iterations $\eta$.) Mu *et al*. [29], Active Subspace [22] and $L_1$-filtering [23] are not compatible with the parameters.

degradation when achieving comparable accuracy with others due to rough approximation on both bases and singular values. Other methods including our FRSVT show similar numbers of iterations and accuracy, but our method has a considerably lower computation time for a single iteration.

Fig. 7 shows iALM behavior over varying over-sampling or power iteration settings. In Fig. 7-(a), varying over-sampling rate $p$ is tested, where we fix the number of power iterations $\eta = 2$ without adopting AP. Without AP, even in the case of $p = 2$, which has been used as our standard experiment setting, iALMs with both FRSVT and RSVD do not converge to the ground truth. This indicates that AP is necessary for a fixed over-sampling rate scheme. Fig. 7-(b) shows the results with the varying number of power iterations $\eta$, where we fix $p = 2$ with AP. The result implies that a single power iteration is sufficient in practice.

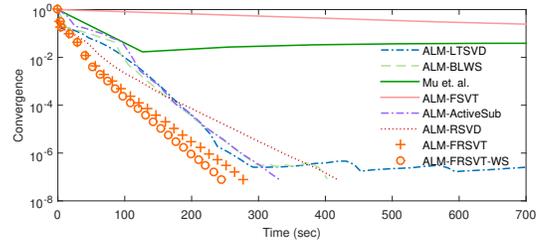

(a) Matlab 2010a

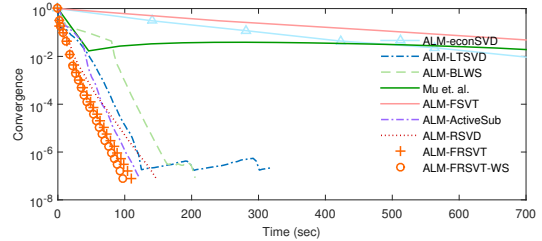

(b) Matlab 2014a

Fig. 6: Convergence time plots of RPCA. X-axis: Elapsed time (sec.), Y-axis: Stopping criterion. Except Mu *et al*. is based on the exact ALM method, all the other methods are based on the inexact ALM. As a representative example, matrices of size $6000 \times 6000$ are used, but the results can be generalized to other size matrices.

### 4.2 Applications

This section shows applications of our FRSVT method to various types of low-rank optimization problems summarized in Table 6, such as affine constrained NNM (Type A), non-convex truncated NNM (Type B) and NNM on tensor structure (Type C). We show them with typical computer vision applications in what follows.

**Type A - Robust Subspace Clustering by Low-Rank Representation (LRR)** Many visual data are often characterized by a mixture of multiple subspaces. One of the recent promising methods is LRR, which effectively performs subspace clustering and noise correction simultaneously. While both noise correction and subspace clustering are known to be challenging, robust LRR has been shown effective even with large corruptions. The robust LRR can be formulated by Type A, which can be efficiently solved by iALM. In this experiment,

| | iALM$_{econSVD}$ | | iALM$_{LTSVD}$ | iALM$_{BLWS}$ | Mu et al. | iALM$_{FSVT}$ | iALM$_{RSVD}$ | iALM$_{FRSVT-RP}$ (Proposed) | Active Sub. | $L_1$ filter |
|---|---|---|---|---|---|---|---|---|---|---|
| | Matlab 2010a | Matlab 2014a | | | | | | | | |
| | | | | | $\|\mathbf{S}\|_0$ | | | | | |
| 10000 × 100 | 99,999 | | 879,415 | 99,999 | 999,999 | 99,999 | 99,999 | 99,999 | 132,666 | 231,604 |
| 2000 × 2000 | 399,997 | | 994,479 | 399,998 | 3,998,401 | 399,997 | 399,993 | 399,997 | 399,991 | 402,932 |
| 4000 × 4000 | 1,599,981 | | 5,607,211 | 1,599,984 | 15,995,318 | 1,599,981 | 1,599,979 | 1,599,981 | 1,599,968 | 1,663,440 |
| 6000 × 6000 | – | 3,599,952 | 35,662,548 | 3,599,966 | 35,990,553 | 3,599,952 | 3,599,955 | 3,599,952 | 3,599,938 | 3,752,782 |
| | | | | | Accuracy | | | | | |
| 10000 × 100 | 7.34e-05 | | 3.39e-02 | 2.66e-04 | 7.65e-01 | 7.34e-05 | 6.75e-05 | 9.33e-05 | 3.00e-02 | 5.16e-02 |
| 2000 × 2000 | 2.11e-07 | | 7.77e-07 | 1.03e-06 | 1.10e+00 | 2.11e-07 | 3.06e-07 | 2.11e-07 | 5.70e-07 | 3.61e-07 |
| 4000 × 4000 | 1.81e-07 | | 4.59e-07 | 4.83e-07 | 1.37e+00 | 1.81e-07 | 2.08e-07 | 1.81e-07 | 3.56e-07 | 2.80e-07 |
| 6000 × 6000 | – | 1.42e-07 | 7.76e-05 | 3.83e-07 | 1.49e+00 | 1.42e-07 | 1.71e-07 | 1.43e-07 | 2.89e-07 | 2.08e-07 |
| | | | | | Total Time | | | | | |
| 10000 × 100 | 3.217 | 1.982 | 1.722 | 1.317 | 125.368 | 2.436 | 0.863 | 0.782 | 1.349 | 3.179 |
| 2000 × 2000 | 2751.910 | 147.055 | 11.075 | 9.808 | 118.798 | 348.132 | 7.928 | 5.123 | 11.163 | 94.300 |
| 4000 × 4000 | 21668.280 | 1068.863 | 69.547 | 53.468 | 759.794 | 2527.141 | 49.027 | 30.274 | 79.471 | 683.159 |
| 6000 × 6000 | – | 3360.379 | 203.051 | 164.752 | 2385.272 | 8238.109 | 147.139 | 88.132 | 246.738 | 1914.431 |
| | | | | | Avg. time for a single iteration | | | | | |
| 10000 × 100 | 0.140 | 0.086 | 0.039 | 0.049 | 1.687 | 0.104 | 0.038 | 0.034 | 0.047 | - |
| 2000 × 2000 | 119.415 | 6.393 | 0.274 | 0.393 | 1.667 | 15.189 | 0.345 | 0.226 | 0.465 | - |
| 4000 × 4000 | 947.908 | 46.481 | 1.726 | 2.185 | 10.664 | 110.006 | 2.132 | 1.348 | 3.311 | - |
| 6000 × 6000 | – | 146.192 | 4.295 | 6.848 | 33.505 | 358.899 | 6.397 | 3.942 | 10.281 | - |
| | | | | | Total no. iteration | | | | | |
| 10000 × 100 | 23 | | 43 | 24 | 73 | 23 | 23 | 23 | 29 | - |
| 2000 × 2000 | 23 | | 41 | 24 | 70 | 23 | 23 | 23 | 24 | - |
| 4000 × 4000 | 23 | | 41 | 24 | 70 | 23 | 23 | 23 | 24 | - |
| 6000 × 6000 | – | 23 | 48 | 24 | 70 | 23 | 23 | 23 | 24 | - |
| | | | | | Speed-up gain against the baseline | | | | | |
| 10000 × 100 | - | 1.62× | 1.87× | 2.44× | 0.03× | 1.32× | 3.73× | 4.11× | 2.38 × | 1.01 × |
| 2000 × 2000 | - | 18.71× | 248.48× | 280.58× | 23.16× | 7.90× | 347.11× | 537.17× | 246.52 × | 29.18 × |
| 4000 × 4000 | - | 20.27× | 311.56× | 405.26× | 28.52× | 8.57× | 441.97× | 715.74× | 272.66 × | 31.72 × |
| 6000 × 6000 | - | - | - | - | - | - | - | - | - | - |

TABLE 5: Quantitative comparison on RPCA. The accuracy is measured by NRMSE with the ground-truth data, and the speed-up gain is computed against the baseline method, iALM$_{econSVD}$ on Matlab 2010a. For Mu et al. [29], we used the implementation and parameters provided by the authors. Since Mu et al. is based on the exact ALM algorithm [18], while other methods are based on the inexact ALM [18], we report the times for a single outer iteration of Mu et al. as 'Avg. time for a single iteration. In addition, the highly stochastic property of the randomly projected nuclear norm makes the algorithm [29] sensitive to data and does not preserve the singular values and sparsity structure of the matrix. This result is consistent with another stochastic method, LTSVD (see and compare $\|\mathbf{S}\|_0$ of LTSVD and Mu et al.). Also, since $L_1$ filtering is incompatible to compare with other methods, the time for a single iteration and the total number of iteration are not reported.

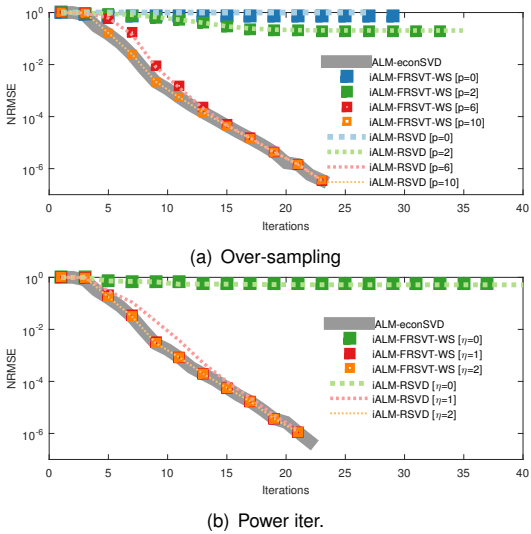

Fig. 7: Varying parameter tests for (a) over-sampling and (b) power iteration in RPCA problem. X-axis: Number of iterations, Y-axis: Normalized root mean square error (NRMSE) of low-rank solution $\mathbf{L}$ as $\frac{\|\mathbf{L}-\mathbf{L}_{GT}\|_F}{\|\mathbf{L}_{GT}\|_F}$.

(a) Over-sampling

(b) Power iter.

| Type | Objective function | Constraint |
|---|---|---|
| A | $\underset{\mathbf{L},\mathbf{S}}{\operatorname{argmin}} \|\mathbf{L}\|_* + \lambda\|\mathbf{S}\|_{2,1}$ | $\mathbf{O} = \mathbf{Z}\mathbf{L} + \mathbf{S}$ |
| B | $\underset{\mathbf{L},\mathbf{S}}{\operatorname{argmin}} \sum_{i=k+1}^{n} \sigma_i(\mathbf{L}) + \lambda\|\mathbf{S}\|_1$ | $\mathbf{O} = \mathbf{L} + \mathbf{S}$ |
| C | $\underset{\mathcal{L},\mathcal{E},\Gamma}{\operatorname{argmin}} \sum_{i=1}^{3} \alpha_i \|\mathcal{L}_{(i)}\|_* + \lambda\|\mathcal{E}\|_1$ | $\mathcal{O} \circ \Gamma = \mathcal{L} + \mathcal{E}$ |

TABLE 6: Examples of NNM related objective functions. (A) Low-rank representation (LRR). (B) RPCA based on the non-convex truncated nuclear norm, where $k$ is the target rank. (C) Low-rank and sparse 3-order tensor decomposition with alignment. Here, $\|\cdot\|_{2,1}$ is $l_{2,1}$ norm, $\{\alpha_i\}$ are balancing parameters among the unfolding matrices $\mathcal{L}_{(i)}$, and $\sum \alpha_i = 1$ is assumed. We refer the basic tensor algebra notations described in [44].

| | Computational Time (s) | | |
|---|---|---|---|
| | LRR+SVD Matlab 2010a | LRR+SVD Matlab 2014a | LRR+Ours |
| Time per Motion | 1.230 | 1.016 | 0.452 |
| Time for 640 Faces | 419.440 | 75.216 | 44.590 |

TABLE 7: Comparisons of the subspace segmentation algorithms on Hopkins155 [40] (Motion) and Extended Yale Database B [16] (Face). Motion Segmentation Errors on the Hopkins155 of both LRR and LRR+Ours are $1.59\%$. The segmentation accuracies (%) on the Yale data are $79.06$ for both LRR and LRR+Ours. These results show the improvement of the computation time with retaining the same accuracy.

we use the same parameter settings and evaluation metrics suggested in [21].

We apply FRSVT to the robust LRR for motion segmentation on Hopkins155 [40] and for face recognition on a part of Extended Yale Database B [16], respectively. All parameters for FRSVT are the same as mentioned in Sec. 2. We use all the 156 sequences in Hopkins155, and the first 10 classes in Extended Yale Database B, in which each class contains 64 face images captured under various illumination conditions – given $42 \times 48$ resized images, we construct a $2016 \times 640$ data matrix by vectorizing each image. While Hopkins155 is only weakly corrupted, Yale data is heavily corrupted by shadows, specularity and noise. As shown in Table 7, our method speeds up LRR without degrading of the accuracy.

**Type B - Semi-Online Weather Artifact Removal** We consider a weather artifacts (e.g., snow or rain) removal problem in a video sequence. Since the weather artifacts have non-deterministic appearance and a sparse yet random distribution in the spatio-temporal domain, we model the artifacts as sparse outlier and the scene as low-rank, while we want to retain moving objects. We apply the low-rank and sparsity decomposition to $n$ frames in a sliding window manner to leave moving objects in the latent images.

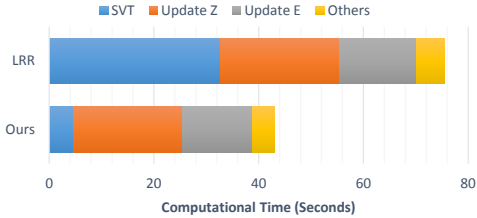

Fig. 8: Computation time comparison on the subspace clustering application [21]. Computation times are measured for face clustering on Extended Yale Database B [16].

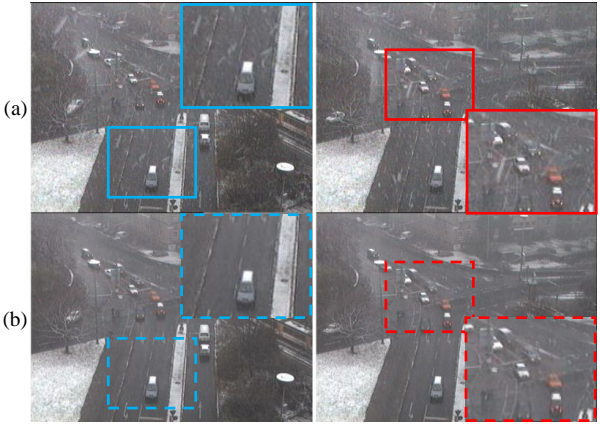

Fig. 9: Qualitative results for the weather artifact removal. (a) Sample input images $\mathbf{O}$. (b) Low-rank images $\mathbf{Z}$.

Unfortunately, since finding low-rank solution by the nuclear norm relies on the high dimensionality of data [4], it could be degraded when only with a small number of frames as reported in Oh *et al*. [30]. With the assumption that the object motions are small within a few frames, we can encourage the low-rank solution to be rank-1 and decompose sparse corruptions by Type B optimization with $k = 1$, where $\mathbf{O}$ is constructed by stacking vectorized $n$ images. It can be effectively solved by alternating Partial SVT (PSVT) [30] and $l_1$ minimization based on iALM. We replace PSVT by FRSVT with the partial thresholding. We transfer the previous basis estimation to the next $n$ frame optimization, so it can be regarded as a semi-online algorithm.

We set the trade-off parameter $\lambda = \frac{1}{\sqrt{\max(m,n)}}$ as suggested by Oh *et al*. [30], and limit the maximum rank-$l$ for computing partial SVD in FRSVT as: $l = 2$ when $n = 3$, and $l = 3$ when $n = 5$ and $10$. We also apply the power iteration twice, *i.e.*, $\eta = 2$. Our algorithm produces $n$ results simultaneously for $n$ images. With an $n = 5$ sliding window, the method based on SVD takes $158.9ms$ per a single channel of a $384 \times 288$ image on Matlab 2014a ($318.3ms$ on Matlab 2010a), while our method only takes $65.5ms$. The qualitative results of our method can be found in Figs. 9 and 10, which show plausible snow removal.

**Type C - Simultaneous Multi-Image Alignment and Rectification** Simultaneous multi-image alignment and rectification problem is Type C optimization suggested by Zhang *et al*. [44] (called SRALT). The method combines the ideas of TILT [45] and RASL [44] on the third-order tensor structure, which exploits the low-rank texture property and nuclear norm-based misalignment error. We show the applicability of FRSVT to NNM on the third-order tensor. We first show that SRALT with FRSVT produces consistent results with the original SRALT on Windows dataset [36] in Fig. 12. Then, we present a new application of SRALT using gait data. We use the parameters $\alpha$ and $\lambda$ in Type-C of Table 6 as: $\lambda = \frac{0.5}{\sqrt{width \cdot height}}$, $\alpha = [0.1, 0.1, 0.8]$ for *Windows*, and $\alpha = [0.15, 0.15, 0.7]$ for the gait data.

Since gait is a biometric signal spanned on not only spatial domain (2D), but also temporal domain, so it is natural to represent the data by a third–order tensor [25]. In gait recognition, the exact alignment of acquired data is frequently assumed, but in real situations, the accuracy of the alignment varies depending on the pivot angle of the camera and locations of moving people. Therefore, the SRALT framework is useful to align gait data making sure that the assumption holds and to find a vertical angle as a preprocessing step. We observe that the human silhouettes have the low-rank texture property, and as observed in Lu *et al*. [25], cyclic gait motions are spanned by a few number of bases.

We conduct the experiment using the Gait Challenge data set [39], called USF Human ID version 1.7. There are 731 gait samples, each with ten sampled gait cycles, and we use the first 300 gait samples. We resize the images into $32 \times 22$, so the size of the input tensor is $\mathcal{O} \in \mathbb{R}^{32 \times 22 \times 3000}$ (3000 unaligned images with the size $22 \times 32$). The synthetic misalignment generation parameters are described in the below. Randomly drawn Euclidean (3 DoF) geometric transformations are applied to each gait image. The parameters are set to: 1) Scale factor: fixed to 1, 2) Angle noise: $N(\frac{10\pi}{180}, (\frac{5\pi}{180})^2)$, 3) Translation noise: $N(0, 0.5^2) + 2 \cdot (U(0,1) - 0.5)$, where $N(\mu, \sigma^2)$ denotes Gaussian distribution with the mean $\mu$ and variance $\sigma^2$, and $U(a,b)$ denotes uniform distribution defined on $[a,b]$.

The set of geometric parameters $\Gamma = \{g_1, \cdots, g_{3000}\}$ (see Type C in Table 6) are defined with a $p$-group parameterization as $g_j \in \mathbb{R}^p$, where 3-DoF parameterization (*i.e.*, $p = 3$) for Euclidean transformation is used in our experiments. Our method is implemented on top of [44] by replacing their SVT method with FRSVT. While the method of [44] converges at the objective value 206.69 and takes about 6 hours 51 min. ($24652s$), our method takes only about 1 hour 45 min. ($6280s$) and achieves an even smaller objective score 206.46. The qualitative and detailed computation time comparisons are summarized in Figs. 13 and 14, respectively.

## 5 DISCUSSIONS

We have presented a fast approximate SVT method that exploits the property of iterative NNM procedures by range propagation and adaptive rank prediction. The approximation error bound shows that our method can produce reliable approximation. The proposed method has been assessed using the problems of affine constrained NNM, non-convex NNM and NNM on tensor structure as well as the original NNM. The empirical evaluations showed the consistent result with the theoretical analysis, and our approach can reduce the computational time of applications that involve low-rank optimization without degrading accuracy and convergence behavior.

For convergence, while a general convergence analysis against various host problems and algorithms is very interesting, it is going to be challenging. RPCA and the Type-A application are convex programs, thus if the algorithms converge with satisfying the respective termination criteria (derived from the convergence criteria [18], [19]), then the algorithms may produce a solution close to global optimal within a small distance (*i.e.*, up to $\epsilon$-optimality). At this point, we can only say that a host algorithm integrated with FRSVT converges to $\epsilon$-optimal, if and only if the problem is convex, the host algorithm converges and FRSVT yields reasonably accurate solutions.

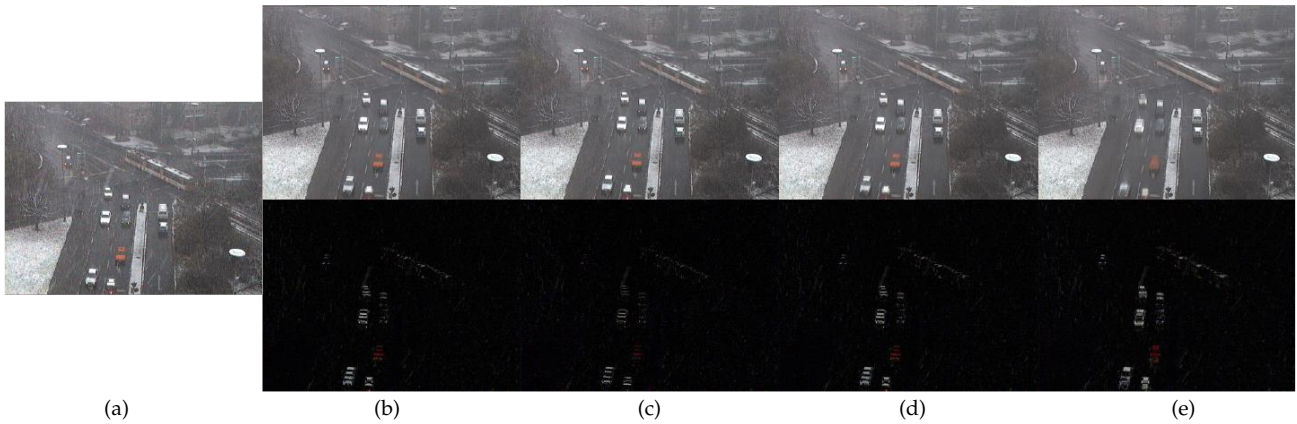

(a)      (b)      (c)      (d)      (e)

Fig. 10: Comparisons according to varying sliding window lengths. (a) Sampled input image. (b) The exact PSVT based result with the sliding window length $5$. (c-e) The PSVT by FRSVT based results with the sliding window length $\{3, 5, 10\}$ respectively. [Top] Low-rank images, $\mathbf{L}$. [Bottom] Absolute of sparse images, $\text{abs}(\mathbf{S})$.

| Computational Time ($ms$) | | | | |
|---|---|---|---|---|
| Sliding Window $n$ | 3 | 5 | 7 | 10 |
| Elapsed Time | 71.5 | 65.5 | 68.2 | 66.2 |

Fig. 11: Computation times of the FRSVT based semi-online weather artifact removal. The elapsed times are measured for a single color channel. Since our algorithm produces $n$ results simultaneously for $n$ images at an execution, we report the time for a single image for a single channel. This results show that, when the number of inputs is small, our weather artifact removal algorithm has linear complexity according to the length of a sliding window, thus the elapsed times for a single image are similar.

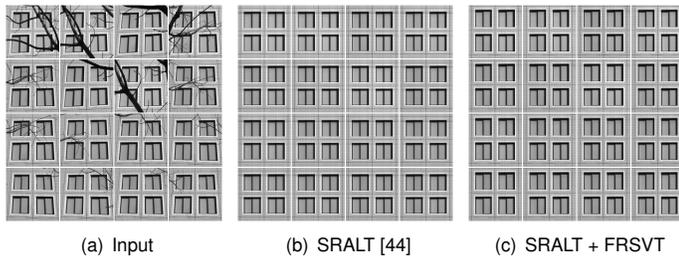

(a) Input     (b) SRALT [44]     (c) SRALT + FRSVT

Fig. 12: Qualitative comparisons for the simultaneously alignment and rectification on *Windows* data [36]. The final solutions produce similar results while the computational speed is enhanced by FRSVT.

The current major computational bottleneck of our method is in the power iteration scheme. We are interested in further reducing the computational complexity by effectively relaxing this computation block in the future.

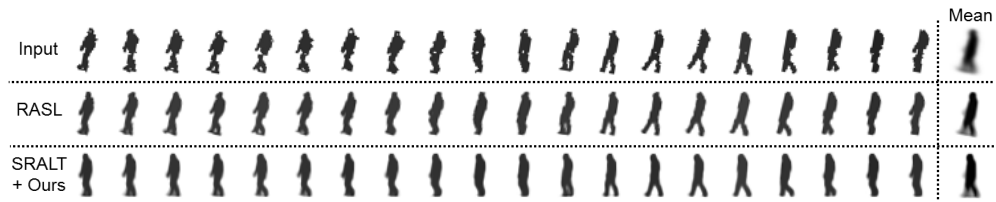

Fig. 13: Qualitative comparison of RASL [36] and SRALT [44] + Our FRSVT. [Top] Samples of input images. [Middle] Results obtained by RASL [36]. [Bottom] Results obtained by SRALT [44] + Our FRSVT. [Right side] Average images. (Top: for all input images, Middle: for RASL, Bottom: for SRALT+Ours). Due to the angular bias shown in (a), RASL aligns samples to be consistent at the biased angle, while the SRALT based method in (c) not only aligns images, but also corrects upright poses.

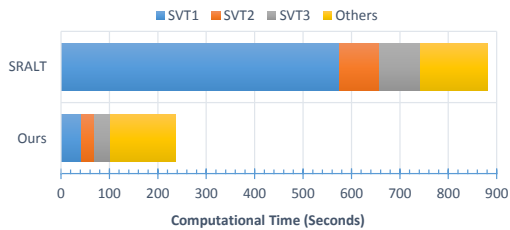

Fig. 14: Computation time comparison on registration and rectification application [44]. The NNM for tensor consists of 3-way SVTs. Computational times are measured for a single outer iteration (almost 100 inner iterations).